\documentclass{article}

\usepackage{arxiv}

\usepackage[utf8]{inputenc} 
\usepackage[T1]{fontenc}    
\usepackage{hyperref}       
\usepackage{url}            
\usepackage{booktabs}       
\usepackage{amsfonts}       
\usepackage{nicefrac}       
\usepackage{microtype}      
\usepackage{lipsum}

\usepackage{graphicx}
\usepackage{float}
\usepackage[labelsep=period]{caption}
\usepackage{enumitem}
\usepackage[utf8]{inputenc}
\usepackage{titlesec}
\usepackage{mymacros}

\title{\textbf{On Intra-Class Variance for Deep Learning of Classifiers}}

\author{
  Rafa\l\ Pilarczyk \\
  \texttt{rpilarcz@gmail.com} \\
   \And
  W\l adys\l aw Skarbek\\
  \texttt{w.skarbek@ire.pw.edu.pl} \\
   \And
   Affiliation\\
    Institute of Radioelectronics and Multimedia Technology \\
    Faculty of Electronics and Information Technology \\ 
    Warsaw University of Technology \\ 
    Nowowiejska 15/19, 00-665 Warszawa, Poland\\
}

\begin{document}
\maketitle

\begin{abstract}
A novel technique for deep learning of image classifiers is presented. The learned CNN models offer better separation of deep features (also known as embedded  vectors) measured by Euclidean proximity and also no deterioration of the classification results
 by class membership probability. The latter feature can be used for enhancing image classifiers having the classes at the model's exploiting stage different from from classes during the training stage. While the  Shannon information of SoftMax probability for target class is extended for mini-batch by the intra-class variance,  the trained network itself is extended by the Hadamard layer with the parameters representing the class centers. Contrary to the existing solutions, this extra neural layer enables interfacing of the training algorithm to the standard stochastic gradient optimizers, e.g. AdaM  algorithm.   Moreover, this approach makes the computed centroids immediately adapting to the updating embedded vectors and finally getting the comparable accuracy in less epochs. 
\end{abstract}

\keywords{
deep  learning of image classifier \and convolutional neural network \and Shanon information measure\and intra-class variance\and image embedding
}

\section{Introduction}

Artificial Neural Networks (ANN) are developed in science and engineering since late 1950s -- the first report in 1957 (Rosenblatt \cite{Rosenblatt57}). Initially, as a algorithmic tool for computer-aided decision making (perceptron), then as a universal mechanism for function approximation  (multilayer perceptron) -- since 1980s when the error backpropagation was published (cf. Werbos' pioneer paper \cite{WerbosP82a} and Rumelhart et al. 
\cite{RumelhartD88a}) using a low dimensional data 
for their classification and regression, and nowadays, since  about 2010, as Deep Neural Networks (DNN) equipped with  specialized operations (e.g. convolutions), operating on large multidimensional signals (also known as tensors) and dozens of processing layers to extract their implicit tensor features. A comprehensive survey of ANN history with a large collection of references can be found in Schmidhuber's article \cite{Schmidhuber14a}.

Deep neural networks algorithms embrace a broad class of  ANN algorithms for data model building in order to solve the hard data classification and regression problems. However, now due to unprecedented progress in computing technology, both tasks can accept digital media signals approaching complexity, the human can brain may deal with.
Moreover, due to the high quality of DNN models, nowadays digital media systems and application significantly improved their performance, comparing to the research status at the turn of the century when the existing media standards (JPEG-x, MPEG-x) were being established.

Discriminant analysis is a procedural concept related to finding, in general hidden features of the analyzed class of objects which make them different from features of other classes of objects being considered. To this goal, usually a separation measure is defined which is used to find a data model supporting the extraction of such discriminant features.  

The classical linear discriminant analysis (LDA) plays the prominent role in pattern recognition systems. Essentially it is based on the Fisher's class separation loss function defined as the ratio of within-class variance to between-class variance (e.g. \cite{Fisher-LDA}, \cite{Skarbek-LDA}). The linear transformation which minimizes the Fisher's discriminant measure is used as the feature extractor which is only a part of end-to-end classifier. Interestingly, the LDA features can be used to recognize objects from classes having no representatives in the training stage of LDA model, e.g. for recognition of persons from a new facial database.

In the case image classifiers defined by DNN  models feature extraction is performed by its CNN (convolutional neural network), which significantly improved state-of-the in computer vision problems like image classification \cite{ren_faster_2017,he_deep_2015,simonyan_very_2014}, image segmentation \cite{badrinarayanan_segnet:_2015,he_mask_2017,chen_deeplab:_2016}, and face modeling \cite{zhang_joint_2016,jourabloo_large-pose_2016}.

In DNN modeling  about separation of feature classes mainly decides the loss function. The most popular objective function is the one which combines the SoftMax function with Shannon information measure. While the former function converts real valued scores into a distribution of probabilities (PDF), the latter one is obtained by using the Kullback-Leibler (KL) divergence measure\footnote{It is known to be equal to cross entropy in this special case.} to this pdf and the crisp membership value for the class the training element belongs to. However, the cross entropy measure is focused on the discrimination of  probability distributions which are computed by DNN on its output -- not on the minimization of intra-class variances relatively to inter-class variances for deep features (also known as embeddings) which are computed by DNN at the end of feature extraction pipeline of network layers\footnote{Usually it is the output of the nonlinear activation following the last convolutional layer in the processing pipeline, preceding full connection layer(s).}. 

{\centering
\begin{figure}[htbp]
\hspace*{-3mm}
\begin{tabular}{cc} 
\includegraphics[width=0.50\linewidth]{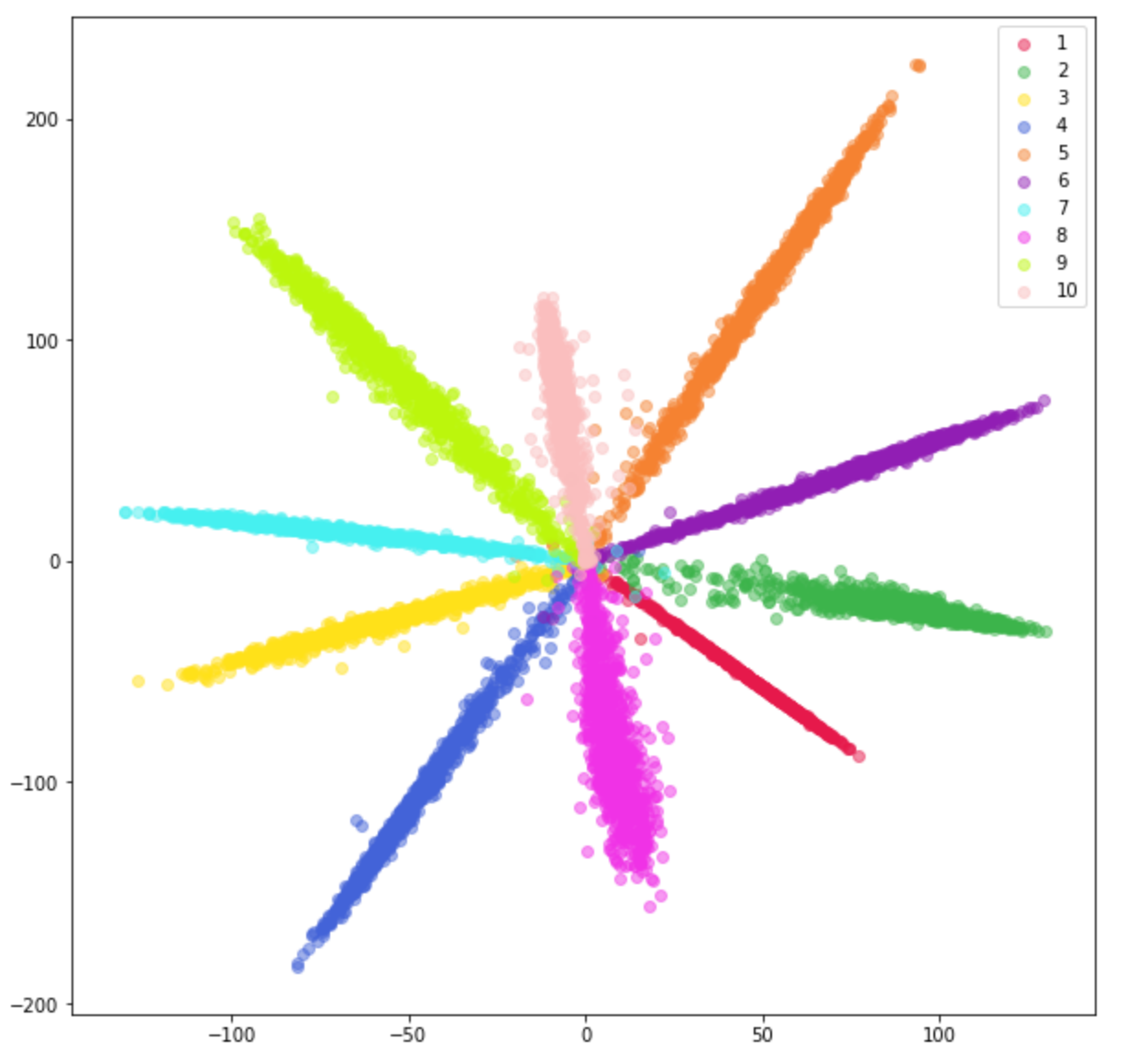} 
 &
 \includegraphics[width=0.48\linewidth]{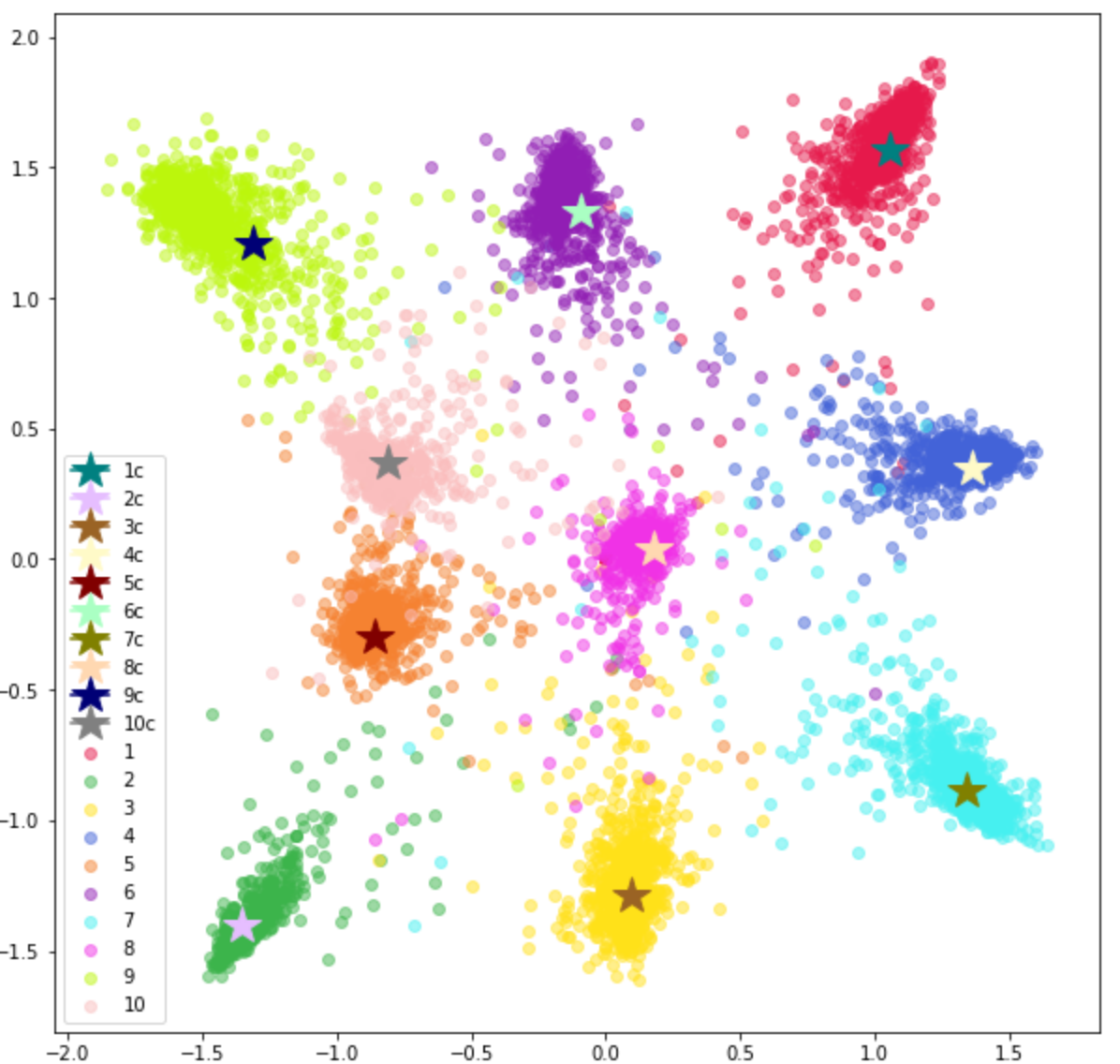} 
\end{tabular} 
\caption{\label{fig:sfm-vs-var}
Deep 2D features (embeddings) for MNIST training data and the loss function: (left) SoftMax followed by the Shanon information function, (right) augmented by intra-class variance term computed  within mini-batch.
}
\end{figure}
}

Embeddings have received immense interest in the research community. The methodology of embeddings has shown a superiority and become an important tool for representation of data for many generic tasks, such as speaker verification, speaker diarization in the audio domain and also for face verification, classification, similarity measurements in the image domain. 

Embeddings modeled as deep neural network and trained on huge amount of data has shown ability to encode different properties of target signal. Enhancements of data embeddings can improve performance the current and potential application of this tool.  Intra-class variances extending probability divergence measures support development of deep features since then we can approximate  the class dependent weighted Euclidean norm by the class independent regular Euclidean norm. It is experimental observation which can be explained by a symmetrical clustering of deep features as  the point cloud representing the embedded vectors of the same class (cf. Figure \ref{fig:sfm-vs-var}).

In the classical optimization to determine variances for the combined loss functions, 
the complete set of training vectors is used for each optimization step. In deep learning practice,  the data sets are huge and therefore stochastic optimization is a reasonable tool and then computing of class separation measure is replaced by its estimation
modified either for each data training sample or for a mini-batch of training data samples. 

In the existing solutions \cite{wen-zhang-li-qiao-16} used for face recognition the exponential weighting 
is performed outside of gradient back-propagation  while the centroid learning 
requires the modification  of standard gradient flow what imposes the  extra complexity 
for the training algorithm. In the paper we show how to re-define the loss functions 
for  both techniques in order to use standard stochastic gradient descent (SGD) optimizers for model training without 
degrading of its performance. 

The research is presented in three sections. In Section \ref{sec:tback} the mathematical concepts related to the discriminant analysis are defined and the basic properties summarized including loss function definitions. In Section \ref{sec:archit} the architectures are presented in symbolic  notation for tensor neural networks. In Section \ref{sec:exper} the experimental results are reported in relevant tables and figures. The paper is closed by the conclusion, the acknowledgment, and the references.

\section{Discriminant analysis -- concepts and basic properties\label{sec:tback}}

In this section mathematical concepts related to the discriminant analysis are summarized: Kullback-Leibler  divergence of probability distribution functions and the data scatter analysis, and the loss functions compared in the research.

\subsection{Information  measure for target class at SoftMax random data  source -- the reference loss function}

Let $x$ be the input tensor of CNN neural network designed for solving of $K$ class problem. The output $p_k\eqd P(x)[k]$ is the probability of $x$ to be in class $k=1,\dots,K.$ We assume that probability distribution $p\eqd P(x)\inv{K}$ is computed from the scores $s\eqd S(x)\inv{K}$ which in turn are processed from  the feature vector $f\eqd F(x)\inv{N}$, i.e. from the embedding of $x$ into $N$ dimensional space \footnote{$e^{s_k}$ is $exp(s_k)$}:
\begin{equation}
x\mapsto F(x)\mapsto s=S(x) \mapsto p=P(x) \lra p_k\eqd \ds \frac{e^{s_k}}{\sum_{i=1}^Ke^{s_i}},\quad k=1,\dots,K
\end{equation}

If the training input $x$ belongs to the class $k$ then the  value of the {\it reference loss function} $\cl{L}_0(x)$ equals to the Shannon information measure\footnote{Information measure of symbol $k$ according Shannon theory of information.} $(-\log P(x)[k])=-\log p_k$ of class $k$ with respect to SoftMax random data source:

\begin{equation}
\text{input } x\ \text{ belongs to class } k \lra \cl{L}_0(x) \eqd -\log P(x)[k] = -\log p_k
\end{equation}

The above definition refers to the single training example $x$.  The Shanon information for the mini-batch is the average information for all mini-batch elements:
\begin{equation}
\cl{L}_0\left(x_{i:i+N_b}\right) = -\frac{1}{N_b}\sum_{j\in[i,i+N_b)}\log(p_{y_j})
\end{equation}

It is really interesting that the definition of $\cl{L}_0$ via Shannon information measure is not used in the literature. Instead we have two other popular definitions: Kullback-Leibler divergence\footnote{By $\bm{1}_k$ we denote the probability distribution $q$ such that $q_k=1.$} $D_{KL}(\bm{1}_k||p)$ and  cross entropy $H(\bm{1}_k,p).$ \footnote{$p_k$ is computed probability distribution for k-th class, $q_k$ - }
It easy to see the equivalence of those definitions:
\begin{equation}
\begin{array}{l}
D_{KL}(q||p) \eqd \dsum_{k=1}^K q_k\log\frac{q_k}{p_k} \lra D_{KL}(\bm{1}_k||p)= -\log p_k = \cl{L}_0(x)\\[5pt]
H(q,p)\eqd  \dsum_{k=1}^K q_k\log\frac{1}{p_k} = H(q)+D_{KL}(q||p) \lra 
\underbrace{H(\bm{1}_k,p))}_0+\underbrace{D_{KL}(\bm{1}_k||p)}_{-\log p_k} = \cl{L}_0(x)
\end{array}
\end{equation}

\subsection{Variability measures}

There are two concepts related to data variability in sets of data samples: variance and scatter. It seems that average squared distance between all pairs of elements in the given set, also known as the {\em set scatter}, is more appealing to daily intuition than the average squared distance to the set's center. Intuitively, whenever we transform data samples before they are assigned to classes, we would like to make their scatter between classes high and within class low.  In this subsection we show that both concepts are mathematically equivalent when computed for the same set of data samples. Moreover the scatter of the sets union embraces their within class and between class variability. We use $\doteq$ as \textit{is equal by definition} term.

\paragraph{Variance and scatter of vectorial data set \\[10pt]}
\hspace*{-5mm}
\begin{tabular}{m{5cm} | m{7cm}}
{\it The variance of a set (sample)} $X$ having discrete probability distribution $p(x)$ normalized to "one", $\sum_{x\in x}p(x)=1$
&
$
\ds\text{var}(X)\eqd\dsum_{x\in X}p(x)\|x-\bar{x}\|^2.
$
\\
\midrule
{\it The scatter of a set (sample)}
 $X$ is the mean of squared Euclidean distance between any pair of sample elements\footnote{For averaging of pair distances we assume independence of selection for elements. Therefore the joint probability equals to the product of probabilities: $p(x,y)=p(x)p(y).$}. 
 &
$
 \text{\small SCATTER}(X) \eqd \frac{1}{2}\dsum_{x\in{X}}\dsum_{y\in{X}}p(x)p(y)\|x-y\|^2
$
\\
\midrule
{\em Property:} Concept of scatter is equivalent to concept of variance.
&
$
 \text{\small SCATTER}(X) = \text{var}(X)
$
\end{tabular}

The variability properties, we discuss here, do not dependent on the name we give to values $p(x)$. The name could be probability or weight -- their positivity and summation to one is important. In context of loss functions to be used for DNN modeling, the weights are interpreted rather as forgetting factors than probabilities.

\paragraph{Class mean and grand mean \\[5pt]}
Let the sample $X$ consists of disjoint subsets $X_k$ including samples drawn from classes\footnote{$[K]$ denotes any set of $K$ indexes, e.g. zero-based indexing $\{0,\dots,K-1\}$ or one-based indexing $\{1,\dots,n\}$.} $k, k\in[K].$ Let  $p_X(x)$ be the probability of $x$ within $X$ while $p_{X_k}(x)$ is the probability of $x$ within $X_k.$ Then\\[10pt]
\hspace*{-3mm}
\begin{tabular}{m{6cm} | m{6cm}}
{\em class mean for sample $X_k$:}  &
$
\bar{x}^k\eqd\dsum_{x\in X_k}p_{X_k}(x)x
$
\\
\midrule
{\em global (grand) mean from sample $X$:} &
$
\bar{x}\eqd\dsum_{x\in X}p_X(x)x
$
\\
\midrule
{\em Property:} The grand mean is the average of class means. &
$
P_k\eqd\dsum_{x\in X_k}p_X(x) \lra \bar{x} = \dsum_{k\in[K]}P_k\bar{x}^k
$
\end{tabular}

\paragraph{Within class variance and covariance\\[10pt]}

Within class (intra-class) variance is important to control separation via attracting elements of the same class. The intra-class covariance matrix can be used to evaluate covariances of within-class embedding errors and then visualize them via initial PCA components, computed after each training epoch.\\[5pt]
  
\hspace*{-5mm}
\begin{tabular}{m{4.5cm} | m{7.5cm}}
{\em Variance} of sample $X_k:$ &
$
\text{var}(X_k)\eqd\dsum_{x\in X_k}p_{X_k}(x)\|x-\bar{x}^k\|^2
$
\\
\midrule
{\em Scatter} of sample $X_k$: &
$
 \text{\small SCATTER}(X_k) \eqd \frac{1}{2}\dsum_{x,y\in X_k}p_{X_k}(x)p_{X_k}(y)\|x-y\|^2
$
\\
\midrule
{\em \un{W}ithin-class variance} is the average of variances for class samples $X_1,\dots,X_K$: &
$
\text{var}_w(X) \eqd \dsum_{k=1}^KP_k\text{var}(X_k)\
$
\\
\midrule
{\em Covariance matrix $R(X_k)$} for sample $X_k$ of  class $k$ &
$
R(X_k)\eqd\dsum_{x\in X_k}p_{X_k}(x)\left(x-\bar{x}^k\right)\tp{\left(x-\bar{x}^k\right)}
$
\\
\midrule
{\em Within-class covariance matrix} is the average of matrix covariances  in classes:&
$
R_w(X) \eqd \dsum_{k\in[K]}P_kR(X_k)
$
\end{tabular}

\paragraph{Within class variance and covariance -- properties\\[10pt]}

The properties show how the within-class variance and covariance matrix are related 
and suggest how covariances of within-class error can be reduced by DNN learning of embedding function without compromising the minimization of within-class variance.\\[5pt]  

\hspace*{-5mm}
\begin{tabular}{m{6cm} | m{6cm}}
{\em Property:} Within-class variance  equals to the trace of within class covariance matrix:
&
$
\text{var}_w(X) = \tr{R_w(X)}
$
\\
\midrule
{\em Property:} The scatter of class sample $X_k$ equals to the variance of class sample $X_k$: 
&
$
 \text{\small SCATTER}(X_k)=\text{var}(X_k),\ k\in[K]
$
\\
\midrule
  {\em Property:}  The within-class scatter, i.e. the average of scatters for class samples equals to the  within-class variance:
    &
$
\dsum_{k\in[K]}P_k\cdot  \text{\small SCATTER}(X_k) = \text{var}_w(X)
$
\\
\midrule
   {\em Property:} The within-class variance is variance for class centered samples:
    &
$
\text{var}_w(X) = \dsum_{x\in X}p_X(x)\|x^{(c)}\|^2,
$\newline
where $x^{(c)}\eqd x-\bar{x}^k,$ if $x\in X_k.$
\\
\midrule
    {\em Property:} The within-class covariance matrix equals to the covariance matrix for class centered samples:
    &
$
R_w(X) = \dsum_{x\in X}p_X(x)x^{(c)}\tp{(x^{(c)})},
$\newline
where $x^{(c)}\eqd x-\bar{x}^k,$ if $x\in X_k.$
\end{tabular}

\subsection{Loss function extension by intra-class variance}

Before adding any new term to the reference loss function $\cl{L}_0$ we should describe the stochastic optimization context. The contemporary DNN optimizers are based on stochastic gradient descent/ascent algorithm (SGD/SGA).  The optimizers themselves are standalone procedures working in the mini-batch mode, i.e.  expecting the gradient $\od{\cl{L}\left(x_{i:i+N_b},y_{i:i+N_b}}{W}\right)$ for the loss function $\cl{L}$ which depends on $N_b$ inputs $x_j$, the $N_b$ outputs $y_j$ drawn from the training sequence, $j\in[i,\dots,I+N_b)$, $i=0,N_b,2N_b,\dots$, and optionally the inner layer(s) outputs(s).  The gradient itself is computed by DNN engine using a type of the AutoGrad algorithm. Its responsibility of the programmer to deliver a differentiable loss function $\cl{L}$ which directly or indirectly depends on network's input data, on the current model (represented by the parameters $W$), and on the desired (by the teacher) network's outputs.

Usually, the programmer computes the {\em instant} loss function for all inputs in the current mini-batch and then makes an aggregation of the results, e.g. averaging, in order to obtain the {\em mini-batch} loss function value. However, there are engines which require the vector of loss function values, the kind of aggregation (reducing) method, and then back-propagate the error into DNN instances, created for all mini-batch elements.

In order to handle learning of centroids $C=[C_1,\dots,C_k]\inm{N}{K}$  we join an additional Hadamard $\bb{H}$ layer\footnote{The Hadamard layer multiplies the input tensor $X$ by the weight tensor $W:$ $H=X*W.$}   with the constant input $\bm{1}_{N\times K}$, parameters matrix $C$, and   the following functionality for the loss term $\cl{L}_{var}$: 
\begin{equation}
\begin{array}{c}
H=\bb{H}(\bm{1}_{N\times K},C) \eqd C, y_{i:i+N_b}\in[K]^{N_b} - \text{ class indexes for the mini-batch} \lra 
\\[5pt]
\cl{L}_{var}(x_{i:i+N_b})\eqd \frac{1}{N_b}\dsum_{j\in[i,i+N_b)}\|x_j-C_{y_j}\|^2 
\end{array}
\end{equation}

The whole loss function $\cl{L}_1\eqd \cl{L}_0+\cl{L}_{var}$. The SGD theory (for instance AdaM  \cite{Kingma18a}) the iterations over the whole batch, i.e. the whole training sequence, lead to the local minimum of $\cl{L}_1$ for certain parameters $W^{\ast}$  in the original part  of DNN and for some $C^{\ast}$ in the extra part. Since $\cl{L}_{var}$ is nonnegative, additive term in $\cl{L}_1$, the local minimum of the whole function implies the local minimum for each. However, the whole batch, i.e. whole training sequence can be divided into $K$  subsequences $J_k$ with elements from the same class. Then over the subsequences $k\in[K]$ we have also a minimum for the variance of class $k$:
\begin{equation}
C_k^{\ast} = \arg\min_{C_k}\left[\dsum_{j\in J_k} \|x_j-C_k\|^2\right]
\end{equation} 

Since $x_j$ depends on the current model $W_j$, the embedded data (deep features) are changing in time, the above reasoning is just intuition -- not really a proof of convergence. 

We get more insight if we notice that minimization of the function $f(C_k)\eqd\sum_{j\in J_k} \|x_j-C_k\|^2$ always leads to the centroid $C_k^{\ast}$ and the minimum value equals to $|J_k|\cdot\text{var}(\{x_j: j\in J_k\})$. Moreover this minimum is global. Since SGD (AdaM  \cite{Kingma18a}) is proved to be convergent with probability one to a local minimum of $\cl{L}_1$, on its part $\cl{L}_{var}$, it is also convergent with probability one to all class centroids giving as the result variances on all class subsequences $J_k, k\in[K].$

Apparently the idea to join classifier loss function with intra-class variance was proposed by Wen et al. \cite{wen-zhang-li-qiao-16}. Their approach to update class centroids was defined outside the error back-propagation  framework. We present the update in another form which gives us conclusion that the update is the exponential, biased weighting of means for embedded vectors $x_j$ from the same class $k\in[1,K]$, occurring in the same mini-batch $i$:
\begin{equation}
\begin{array}{c}
 J_k'\eqd\{j\in[i,i+N_b): y_j=k\},\ n_k\eqd |J_k'|,\ n_k\neq 0 \lra\\[5pt]
  C_k := (1-\alpha)\cdot C_k+\ds\frac{\alpha}{n_k}\cdot\sum_{j\in J_k'}x_j
\end{array}
\end{equation}

There is also another possibility to make weighting of embedded vectors and centroids one by one without averaging them in mini-batches ($n_k$ term) (\cite{Kowalski-Naruniec18}):
\begin{equation}
 j\in[i,i+N_b),\ y_j=k \lra  C_k := (1-\alpha)\cdot C_k+\alpha\cdot x_j
\end{equation}
 This is also biased, exponential weighting of embedded vectors. However the embeddings defined by function $F$ are modified after each batch. Moreover, having the centroids iterated outside of the gradient framework we cannot use for them the conclusions on stochastic convergence to local minimum by SGD optimization method (e.g. AdaM algorithm \cite{Kingma18a}). Our updates for centroids are integrated with SGD framework with learning factors modified according embedding data evolution what is the main reason of provable stochastic convergence.
 
 Though in this paper we do not present the loss function option with the subtracted term for inter-class variance, the concept of nonlinear discriminant analysis in principle is possible to produce yet another embeddings for various applications. In such extension the SGD optimizer is also doing the main work for updates not only for class centroids but also it updates the grand mean centroid while keeping bounded the values of the total variance for all embedded vectors.

\section{CNN architectures in symbolic notation\label{sec:archit}}

\subsection{Elements of symbolic notation}

The symbolic notation for DNN architectures follows the definitions of Symbolic Tensor Neural Networks presented for the first time in the tutorial \cite{Skarbek-stnn} of the second author. The selected rules of this notation are given in the table below:\\[10pt]

\begin{tabular}{m{1cm} | m{11cm}}
Symbol & Description\\
\toprule\toprule
\xin{a}{b}{c} & Input layer of name $(a)$ with tensor of signal axes $(b)$ and feature axis $(c)$, e.g.\xin{yx}{1}{image}\\
\midrule
\xconv{a}{b}{c}{}{d} & Convolutional layer with $(b)$ kernels of dimensions and strides (if any) $(a)$ having options $(c)$ followed by activation(s) $(d)$, e.g.\xconv{5}{64}{p}{}{br} where there are $64$ convolutions with kernels of size $5$ at each signal axis, zero padding option is required, and the batch normalization and ReLU activation is performed, afterwards.\\
\midrule
\xpool{a}{}{b}{}{} & Pooling layer which aggregates feature values in blocks of size $(a)$ with optional striding $(a)$ according to a technique $(b)$, e.g.\xpool{2}{}{m}{}{} -- max pooling in blocks of size $2$ at each signal axis is performed.\\
\midrule
\xdense{}{a}{}{}{b} & Full connection layer with $(a)$ output features followed by activation(s) $(b)$, e.g.\xdense{}{4096}{}{}{br} where feature vector with $4096$ components is computed by linear operation on layer's tensor input. $b$ in this example is batch-normalization, $r$ - ReLU activation function\\
\midrule
\xdrop{a} & The layer drop-outs randomly $(a)$ percents of weights in the next layer. It is only used at training, e.g.\xdrop{50} makes zeroing for $50\%$ of weights. 
\end{tabular}

\subsection{Face descriptor -- CNN architecture with centroid layer}

Descriptor extraction for face recognition fits to the first above category. The solution proposed by Jacek Naruniec and Marek Kowalski of Warsaw University of Technology (KN-FR \cite{Kowalski-Naruniec18}) is a combination of convolutional neural network with loss function aggregating SoftMax loss with intra-class variance, proposed by Wen, Zhang et al. in \cite{wen-zhang-li-qiao-16}: {\it A Discriminative Feature Learning Approach for Deep Face Recognition}.

In the architecture described below we added the extra centroid layer which is referenced from the loss function -- the term for intra-class variance. Therefore the normalized embeddings into the Euclidean space $\bb{R}^{1024}$ are different then the embeddings in \cite{wen-zhang-li-qiao-16} and \cite{Kowalski-Naruniec18}.

\begin{center}
\doublebox{
\begin{tabular}{l}
\xin{yx}{1}{image}
\xconv{5}{64}{p}{}{br}
\xconv{5}{64}{p}{}{br}
\xconv{5}{64}{p}{}{br}
\xpool{2}{}{m}{}{}
\xconv{5}{128}{p}{}{br}
\xconv{5}{128}{p}{}{br}
\xconv{5}{128}{p}{}{br}
\xpool{2}{}{m}{}{}
\\[5pt]
\xconv{5}{256}{p}{}{br}
\xconv{5}{256}{p}{}{br}
\xconv{5}{256}{p}{}{br}
\xconv{5}{256}{p}{}{br}
\xpool{2}{}{m}{}{}
\xconv{5}{512}{p}{}{br}
\xconv{5}{512}{p}{}{br}
\xconv{5}{512}{p}{}{br}
\xconv{5}{512}{p}{}{br}
\xpool{2}{}{m}{}{}
\\[5pt]
\xconv{5}{512}{p}{}{br}
\xconv{5}{512}{p}{}{br}
\xconv{5}{512}{p}{}{br}
\xconv{5}{512}{p}{}{br}
\xpool{2}{}{m}{}{}
\xdense{}{4096}{}{}{br}
\xdense{}{1024}{}{}{}\xtolabel{x}
\\[3pt]
\xmath{x/\|x\|}
\xtolabel{norm}
\xdense{}{P}{}{}{}
\xtolabel{scores}
\quad\quad\quad
\xfromlabel{ones}\xmath{\bm{1}\cdot C}\xtolabel{centers}
\\
\midrule\midrule
\xbound{knFaRec}{}{
\begin{array}{rcl}
image &:=& 112_{yx};\\ 
optima &:=& \left[loss,MomentumSGD,\right.\\
&&\left.ShannonInfo+IntraClassVar\right]
\end{array}
}
\end{tabular}
}
\end{center}

Note that embedded vectors are normalized here in order to get the unit length.

\subsection{MNIST descriptors -- CNN architecture with centroid layer}

In order to illustrate intuitively the concept of embeddings controlled by learned centroids we consider the generic MNIST problem for handwritten digits recognition and Fashion MNIST recognition for fashion images. For both problems the following CNN architecture is trained.

\begin{center}
\doublebox{
\begin{tabular}{l}
\xin{yx}{}{image}
\xconv{3}{16}{}{}{r}
\xconv{3}{32}{}{}{r}
\xpool{2}{}{m}{}{}
\xdrop{50}
\xconv{3}{64}{}{}{r}
\xconv{3}{64}{}{}{r}
\xpool{2}{}{m}{}{}
\xdrop{50}
\xdense{}{n}{}{}{r}
\xtolabel{x}\\[5pt]
\xfromlabel{x}
\xdense{}{10}{}{}{}
\xtolabel{scores}
\quad\quad\quad
\xfromlabel{ones}\xmath{\bm{1}\cdot C}\xtolabel{centers}
\\
\midrule\midrule
\xbound{embed-n}{}{
\begin{array}{l}
image:= 28_{yx};\\
optima := [loss, AdaM, ShannonInfo+IntraClassVar]
\end{array}
}
\end{tabular}
}
\end{center}

The feature (embedded) vector $x\inv{n}$ in this study is normalized or not-normalized.  In our experiments, the embedding space dimensionality is low and equals to $n=2, 4.$ 

\begin{figure}[htbp] 
\hspace*{-5mm}
\begin{tabular}{cccc}
 \includegraphics[width=0.23\linewidth]{pics/2_dim_pictures/mnist/nonnormalized_2_softmax.png}
 &
 \includegraphics[width=0.23\linewidth]{pics/2_dim_pictures/mnist/nonnormalized_2_intra.png} 
 &
 \includegraphics[width=0.23\linewidth] {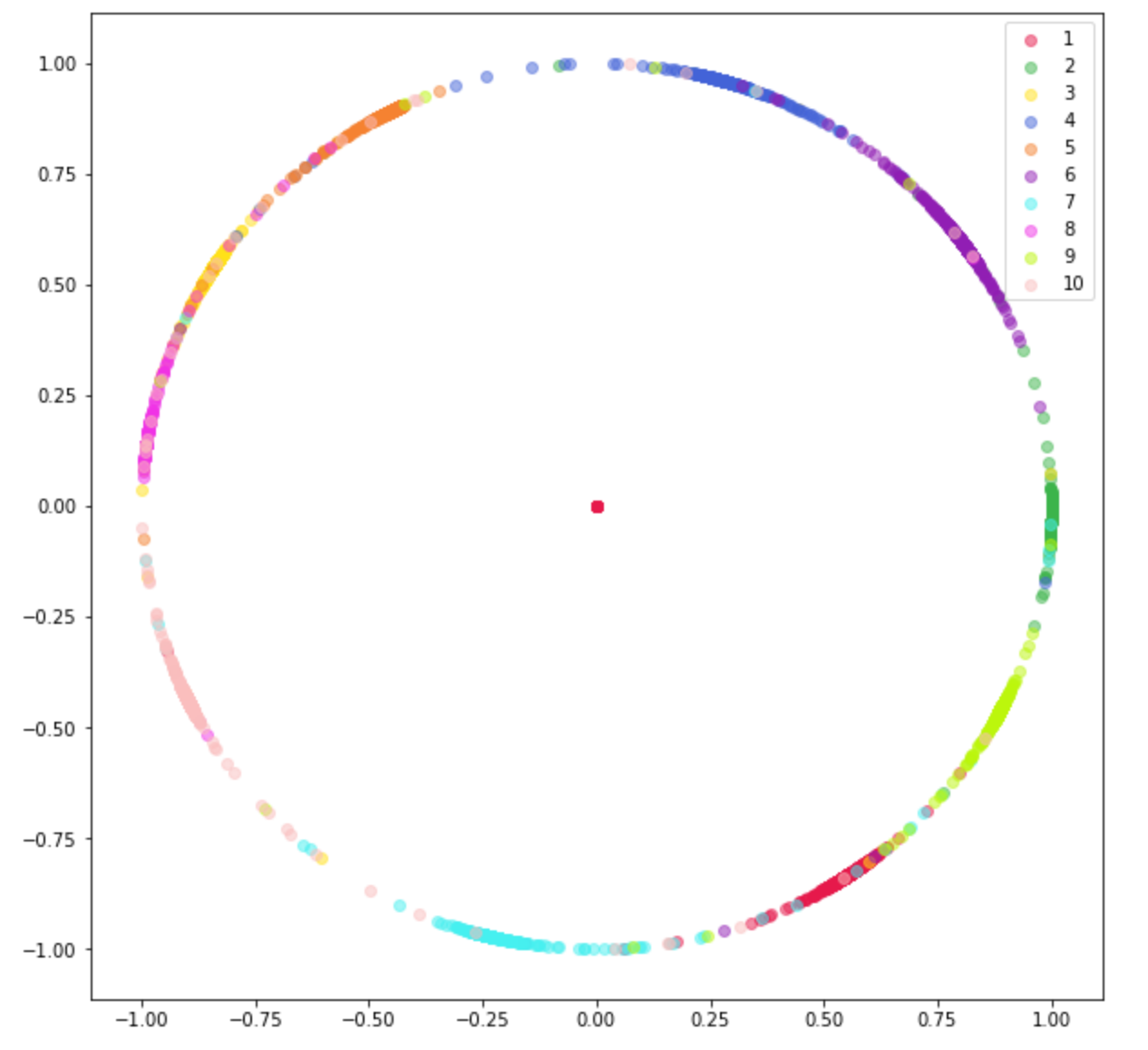}
&
\includegraphics[width=0.23\linewidth] {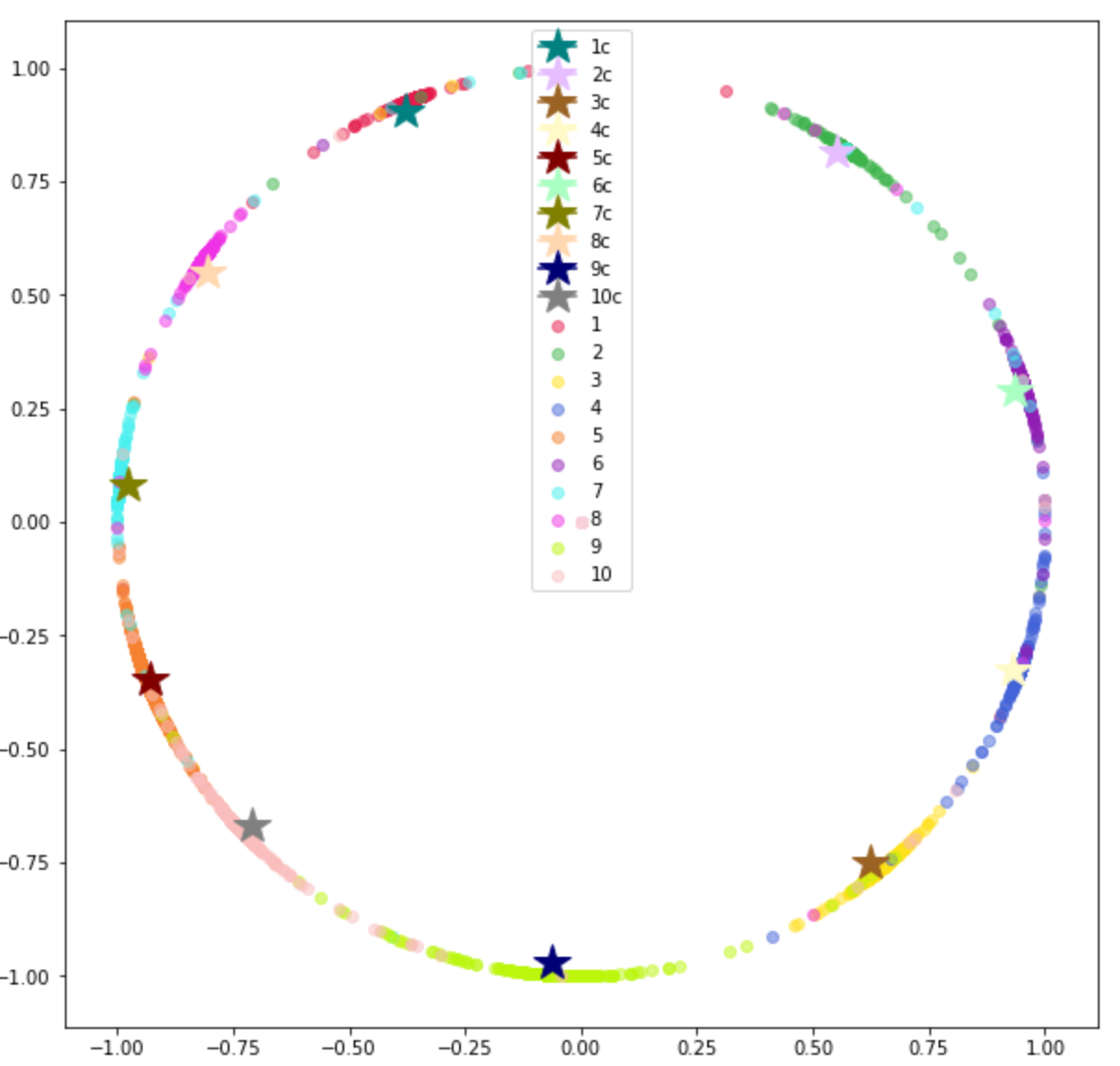}
\end{tabular}
\caption{\label{mnist-2}MNIST 2-dim embeddings visualization.}  
 \end{figure}

\begin{figure}[htbp] 
\hspace*{-5mm}
\begin{tabular}{cccc}
 \includegraphics[width=0.23\linewidth]{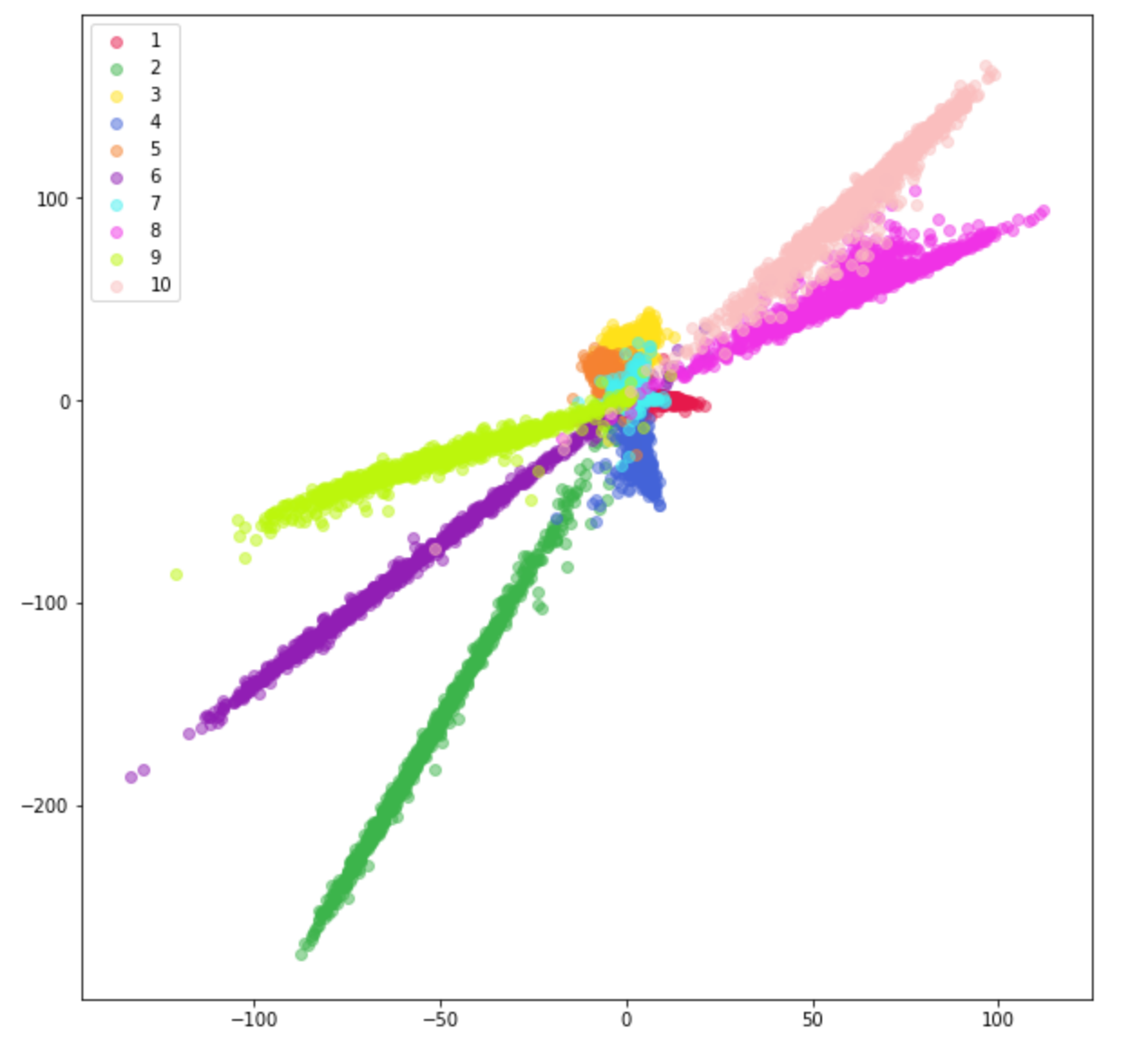}
 &
 \includegraphics[width=0.22\linewidth]{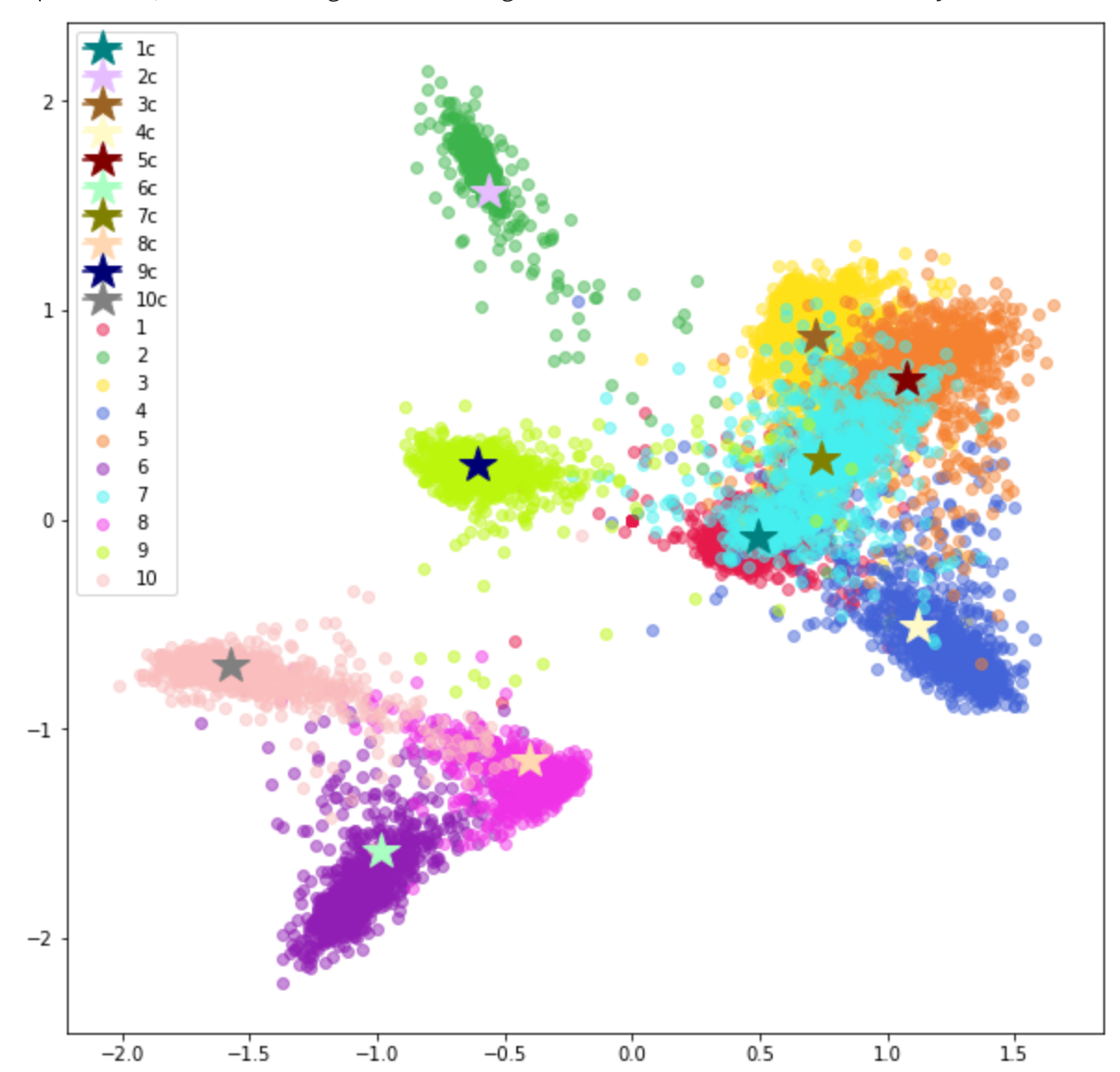} 
 &
 \includegraphics[width=0.23\linewidth] {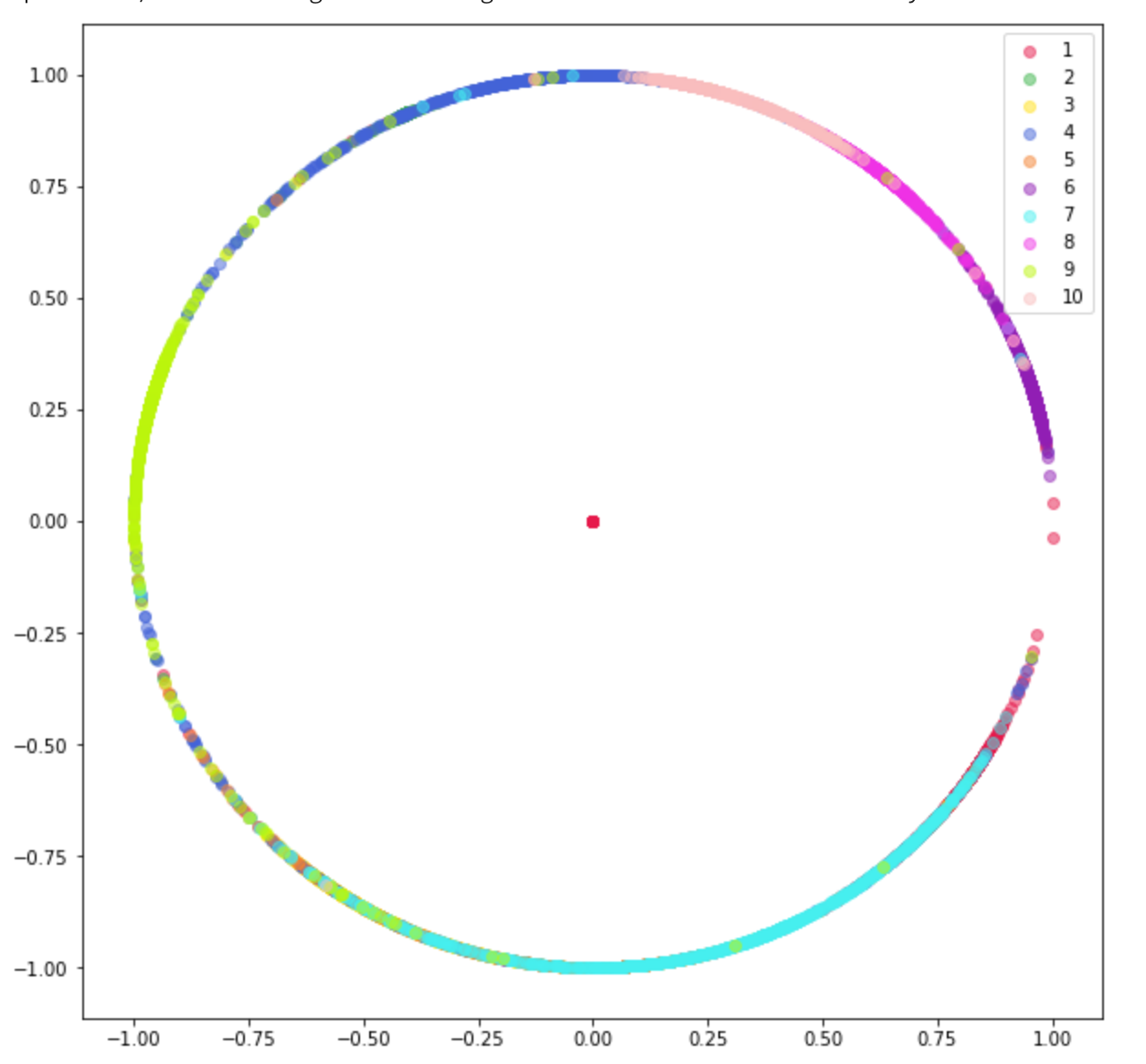}
&
\includegraphics[width=0.23\linewidth] {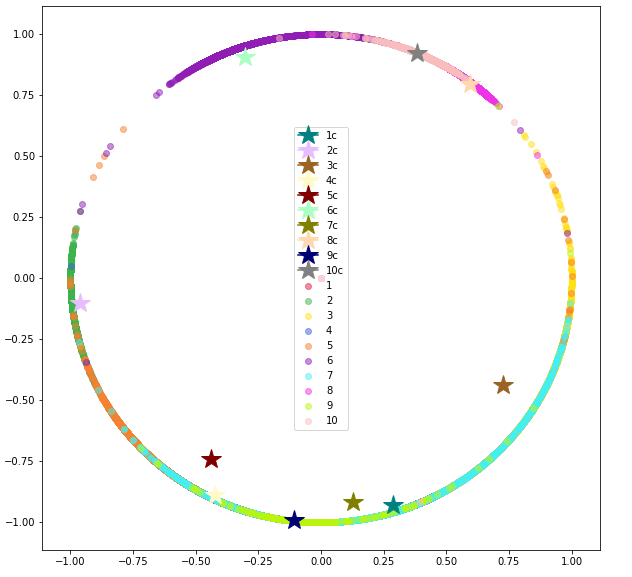}
\end{tabular}
 \caption{\label{fashion-2}Fashion MNIST 2-dim embeddings visualization .}
\end{figure}

\section{Experiments \label{sec:exper}}

Our experiments are carried out on convolutional neural network, changing the size of the layers used to extract embeddings. MNIST \cite{mnistlecun} and Fashion-MNIST \cite{xiao_fashion-mnist:_2017} datasets are used. The comparative analysis is aimed at indicating potential factors in which the loss function should be modified in order to achieve optimal results for embeddings similarity. 

\begin{figure}[htbp] 
\hspace*{-5mm}
\begin{tabular}{cccc}
    \includegraphics[width=0.23\linewidth] {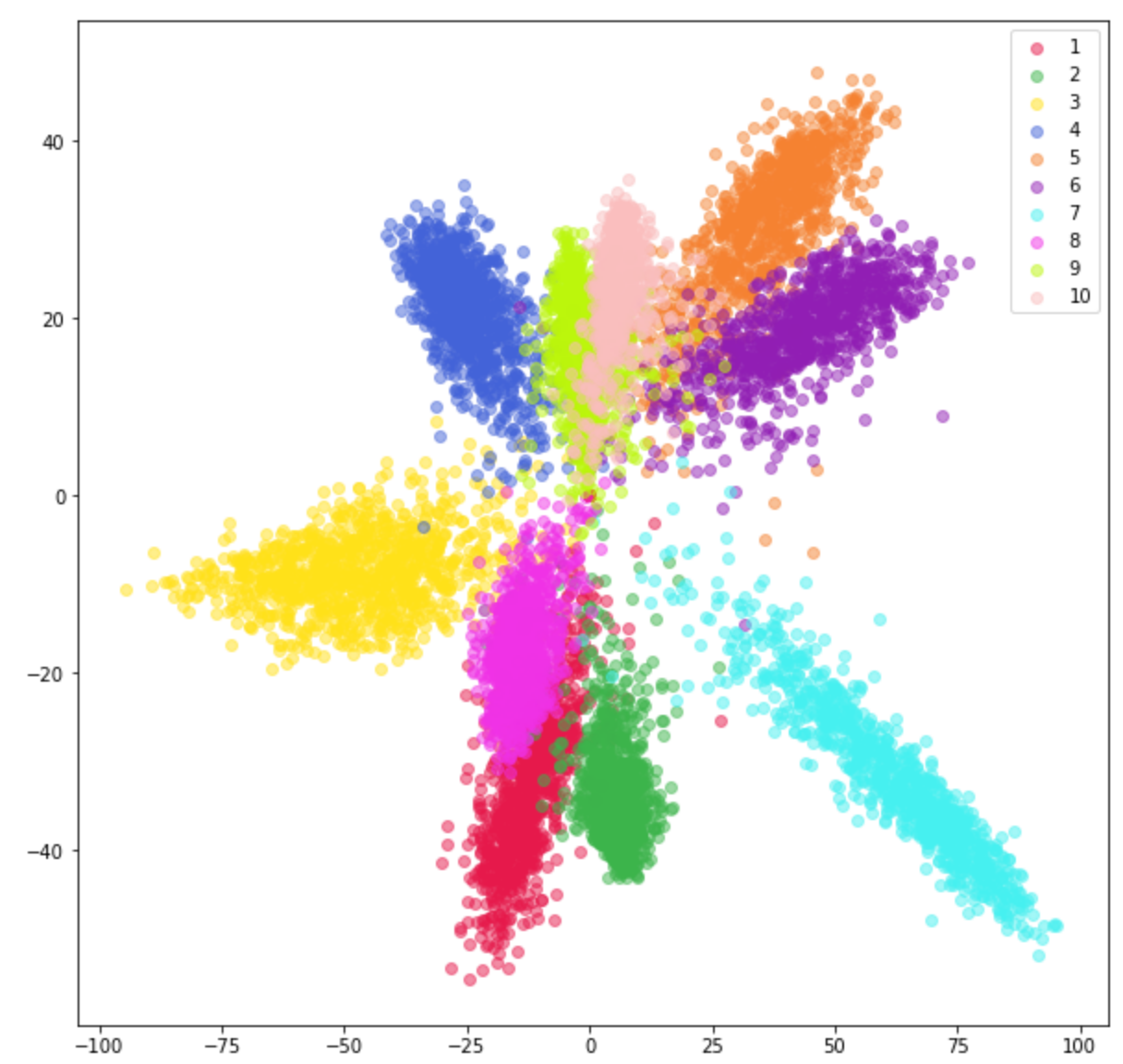} 
&    \includegraphics[width=0.23\linewidth] {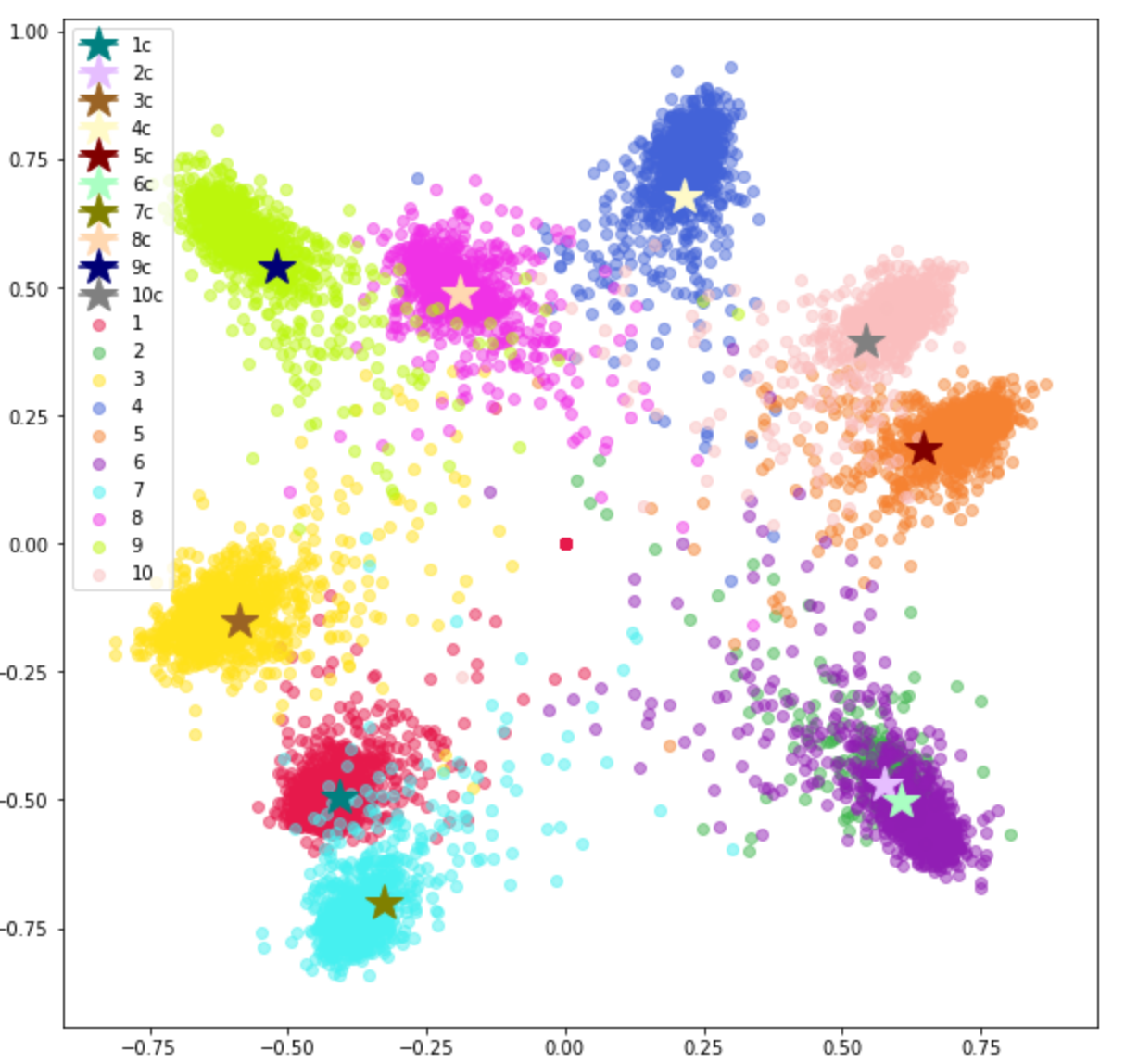} 
&    \includegraphics[width=0.23\linewidth] {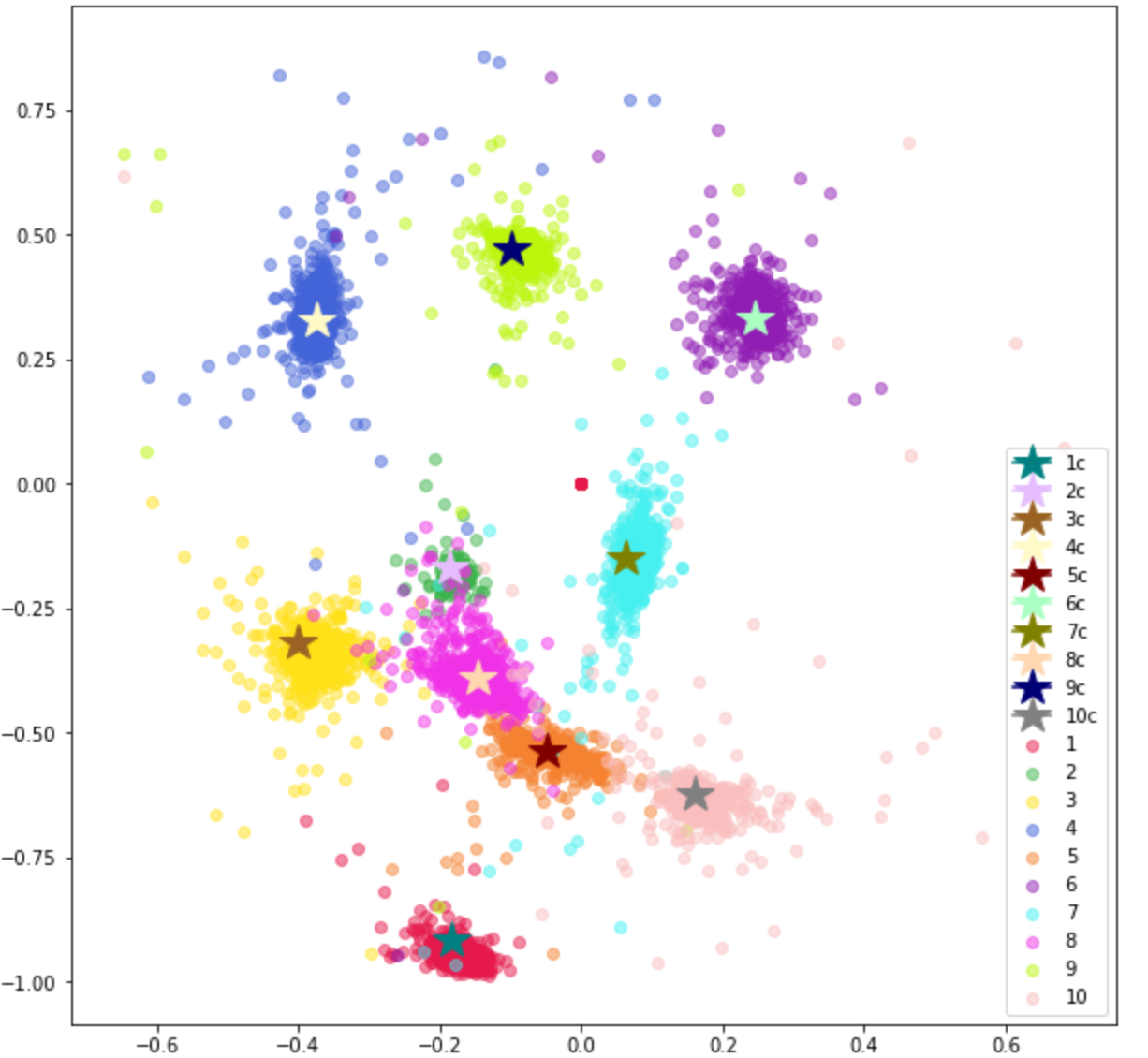}
&    \includegraphics[width=0.23\linewidth] {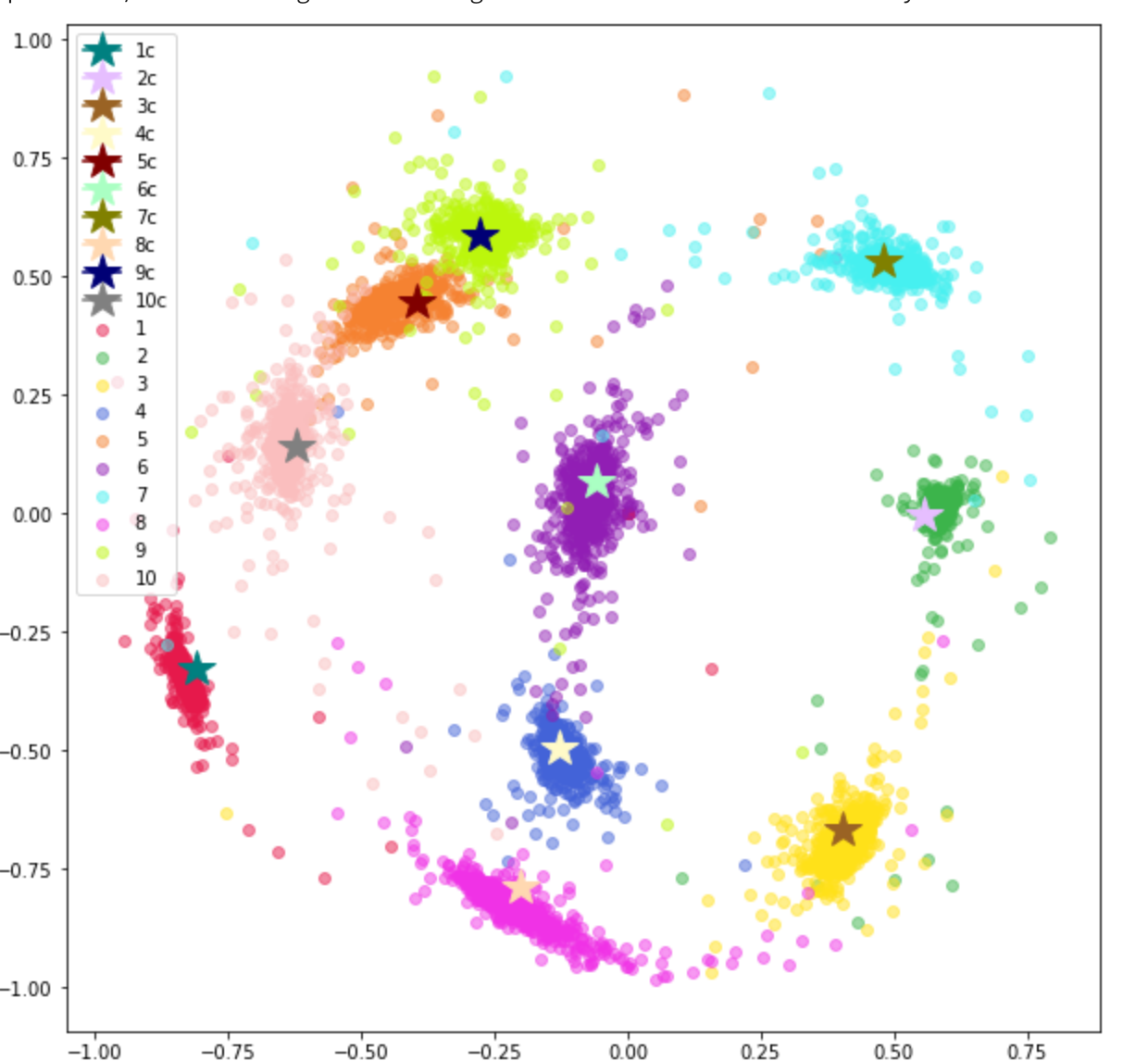}
\end{tabular}
  \caption{ \label{mnist-4}MNIST 4-dim embeddings visualization using 2-component  PCA.}
\end{figure}    

Most of prior works focus on large-scale problems, where training hyperparameters can have as significant impact as loss function selection, architecture or weights initialization. Therefore in this paper we focus on small-scale problem followed by many experiments with the same initial parameters to see where large-scale problems like face recognition can have possible bottlenecks and how to improve it.

\begin{figure}[htbp] 
\hspace*{-5mm}
\begin{tabular}{cccc}
    \includegraphics[width=0.23\linewidth] {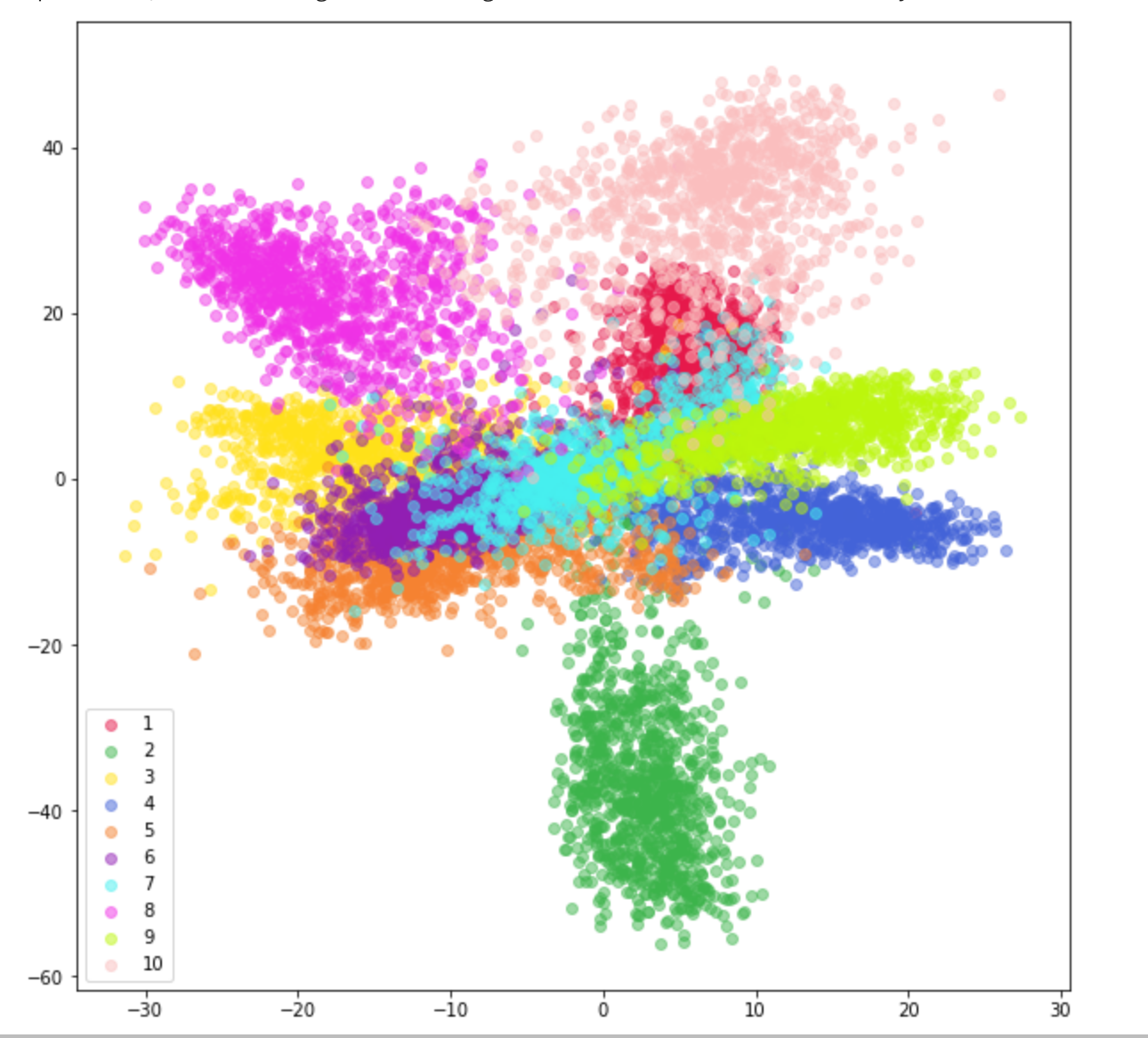} 
&    \includegraphics[width=0.23\linewidth] {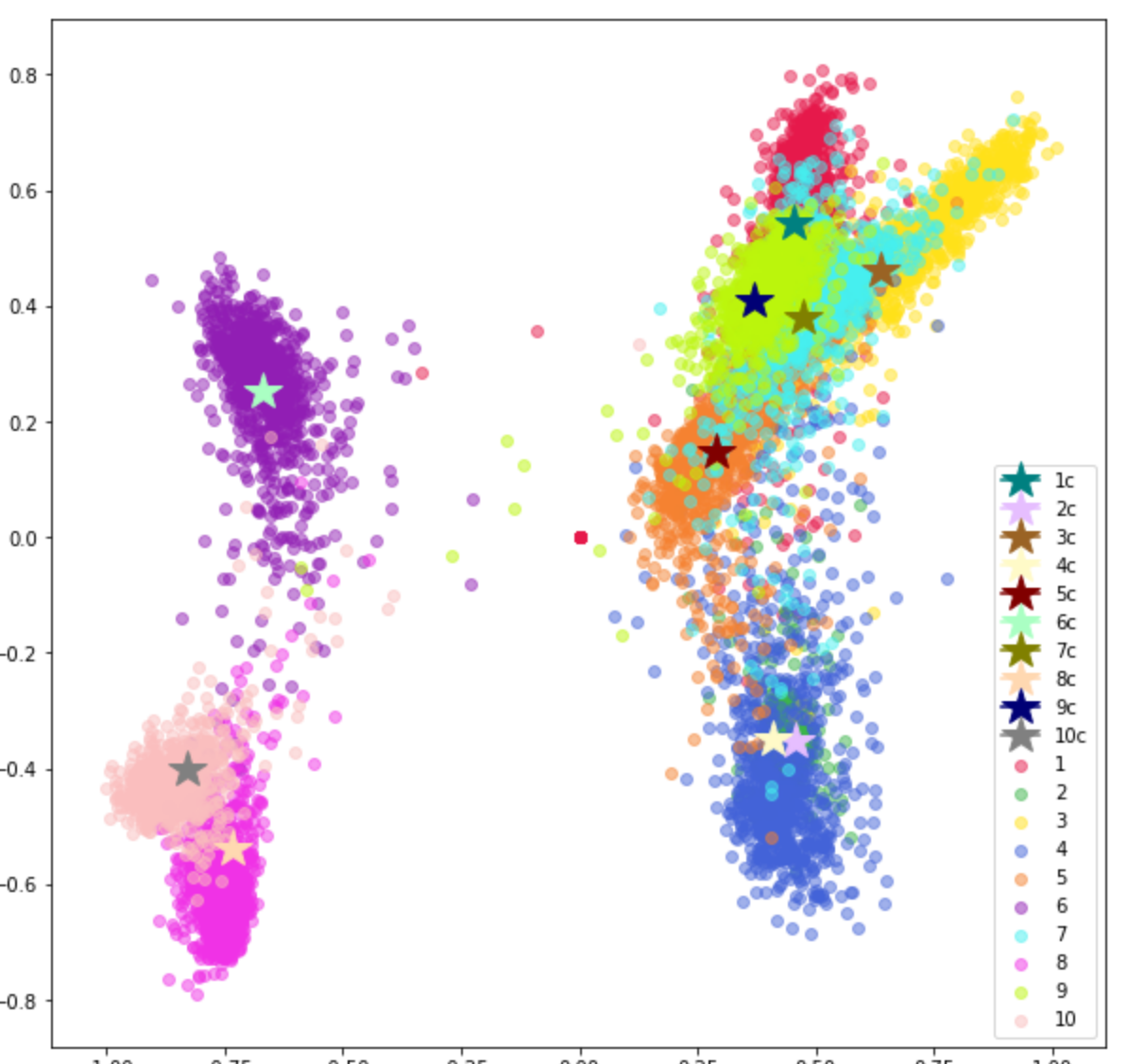} 
 &   \includegraphics[width=0.23\linewidth] {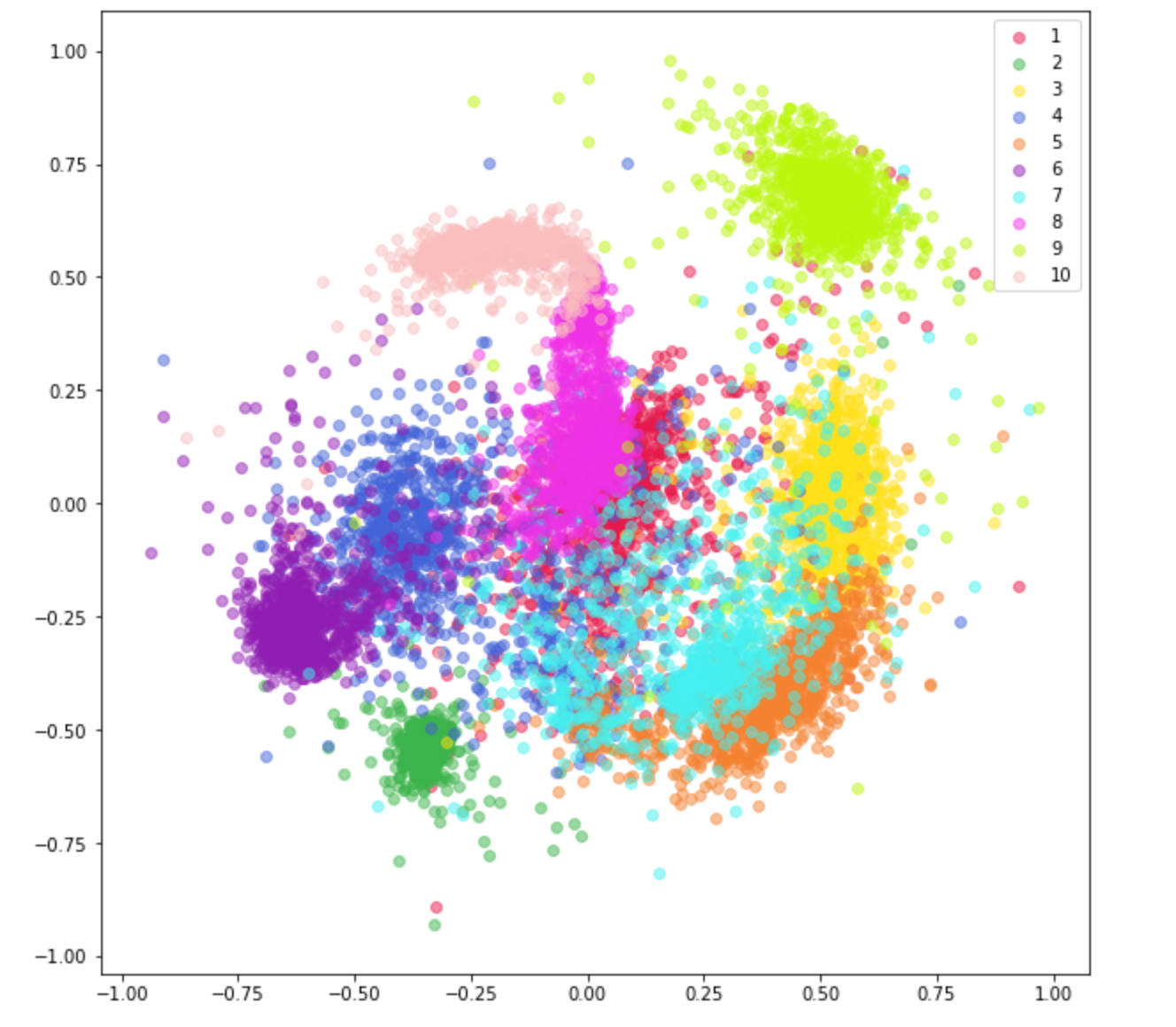}
&    \includegraphics[width=0.23\linewidth] {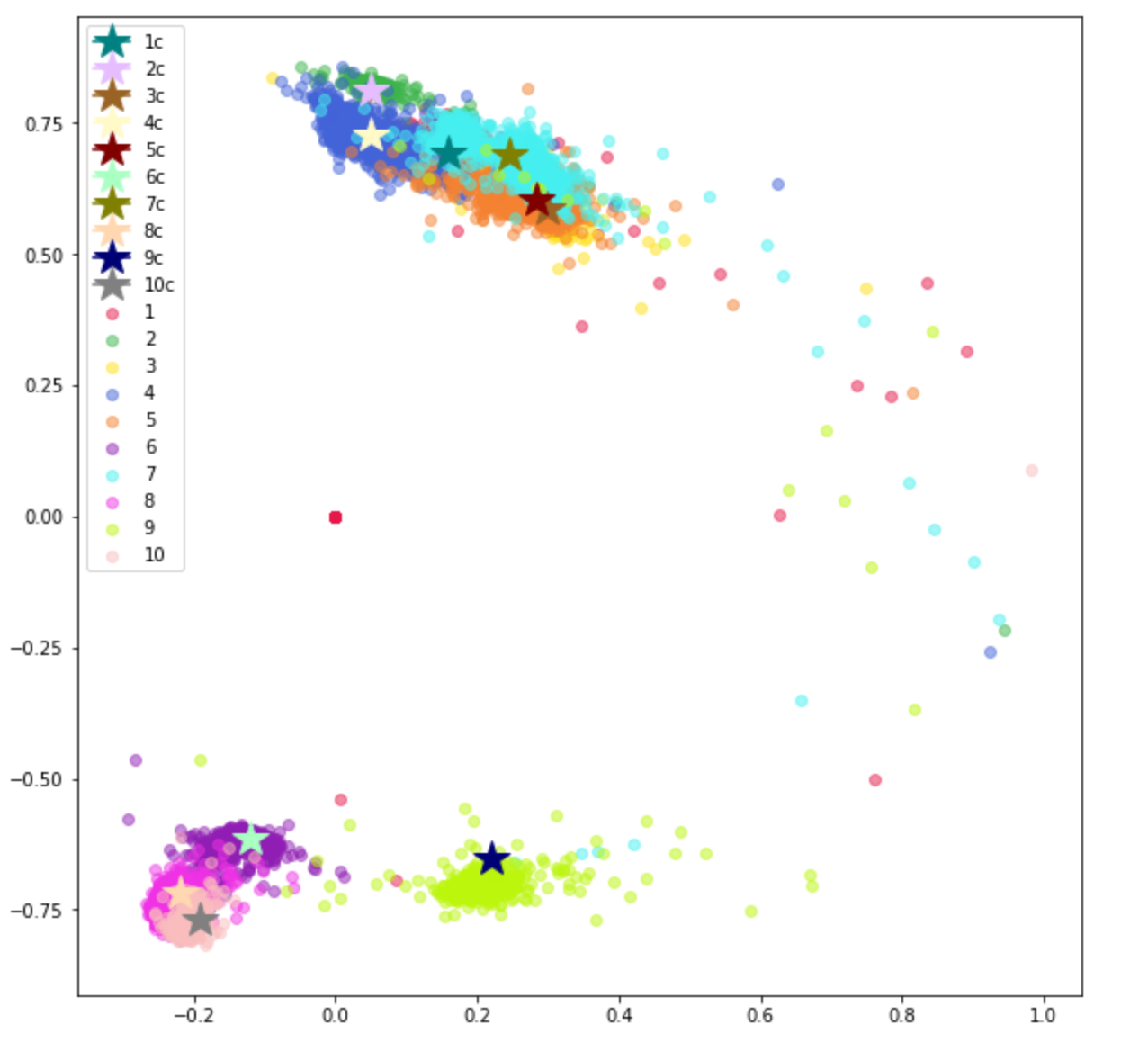}
  \end{tabular}
  \caption{\label{fashion-4}Fashion MNIST 4-dim embeddings visualization using 2-component PCA.}
\end{figure}

\subsection{Implementation details}

For training we use AdaM optimizer with learning rate 0.001, mini-batch size of 256. Each network is trained over 20 epochs without any additional learning rate drop. We change the size of bottleneck embedding layer and also optionally apply normalization of features in experiments. For visualization purposes of more than 2 dimensional space we use 2-component PCA decomposition.

\begin{figure}[htbp] 
  \begin{minipage}[b]{0.19\linewidth}
    \centering
    \includegraphics[width=1.0\linewidth]{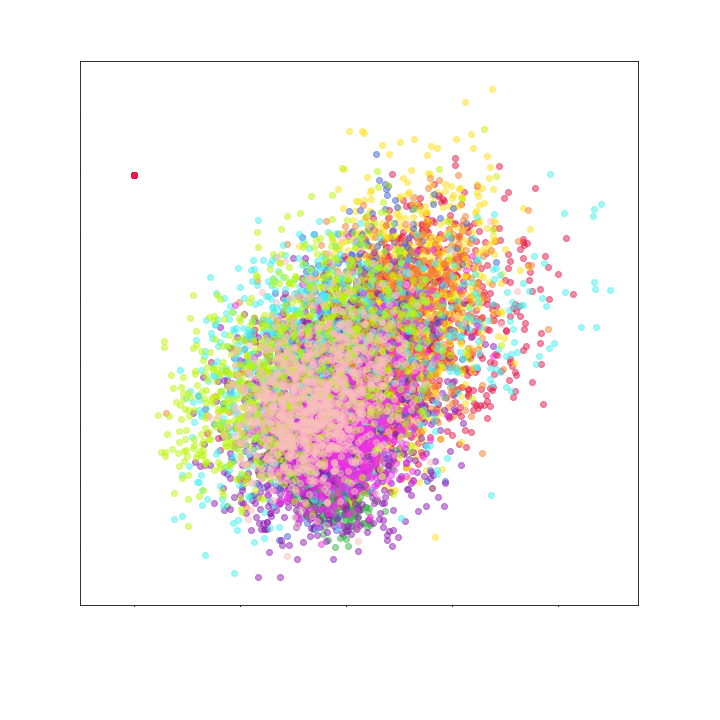}
  \end{minipage}
  \begin{minipage}[b]{0.19\linewidth}
    \centering
    \includegraphics[width=1.0\linewidth]{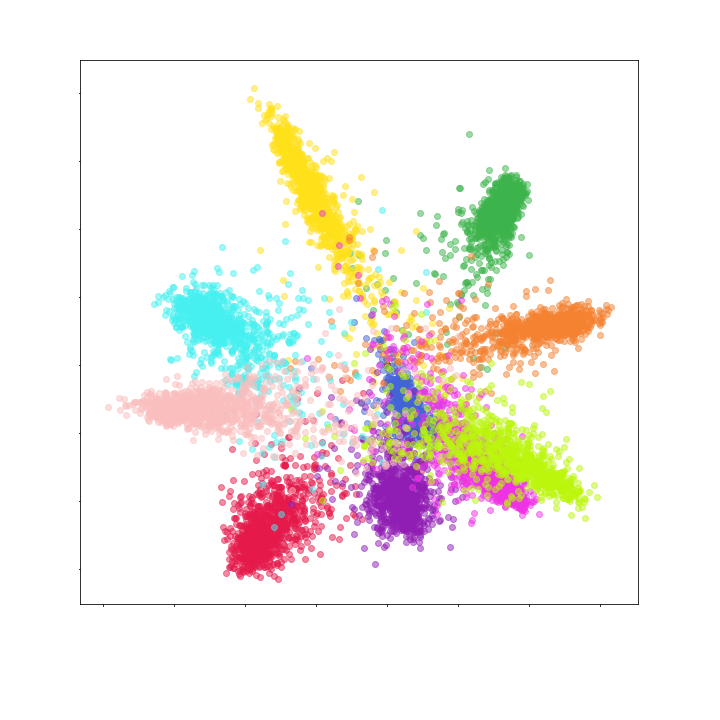}
  \end{minipage} 
  \begin{minipage}[b]{0.19\linewidth}
    \centering
    \includegraphics[width=1.0\linewidth]{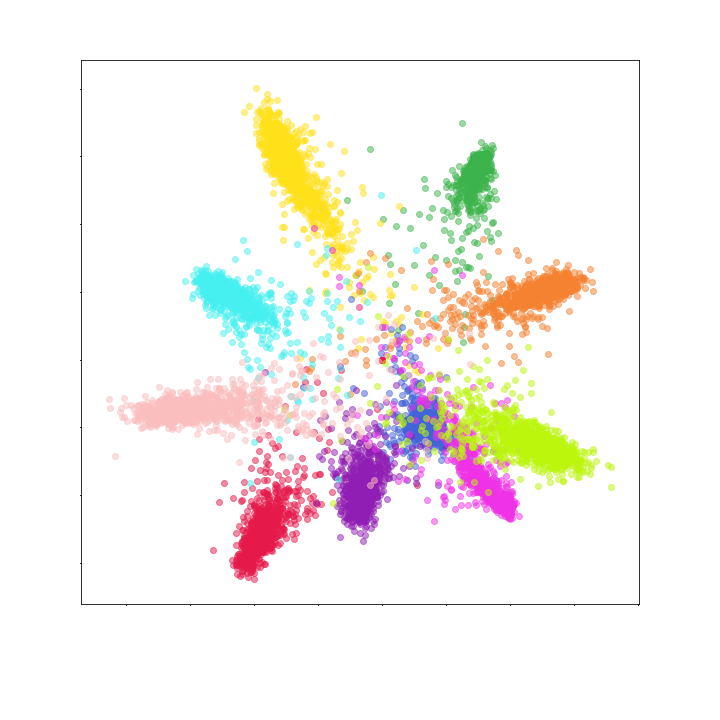}
  \end{minipage}
  \begin{minipage}[b]{0.19\linewidth}
    \centering
    \includegraphics[width=1.0\linewidth]{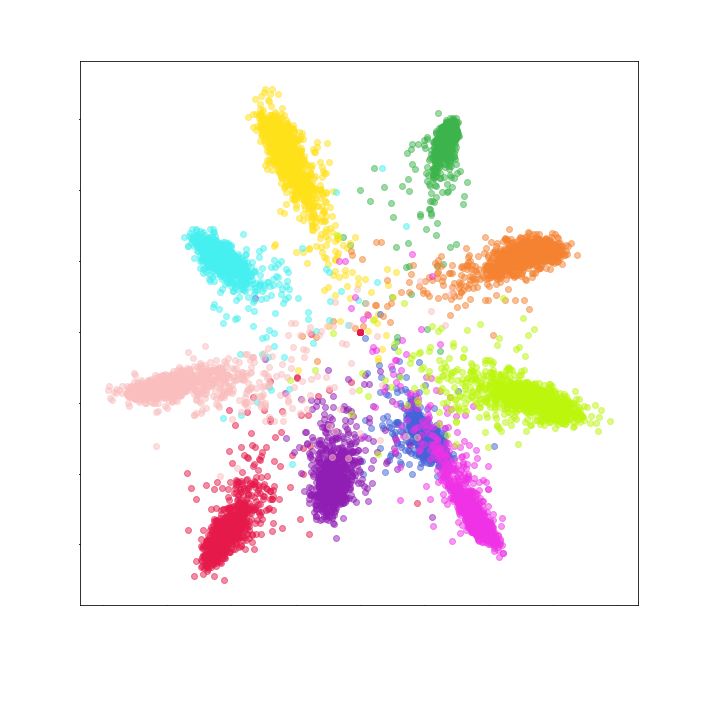}
  \end{minipage}   
  \begin{minipage}[b]{0.19\linewidth}
    \centering
    \includegraphics[width=1.0\linewidth]{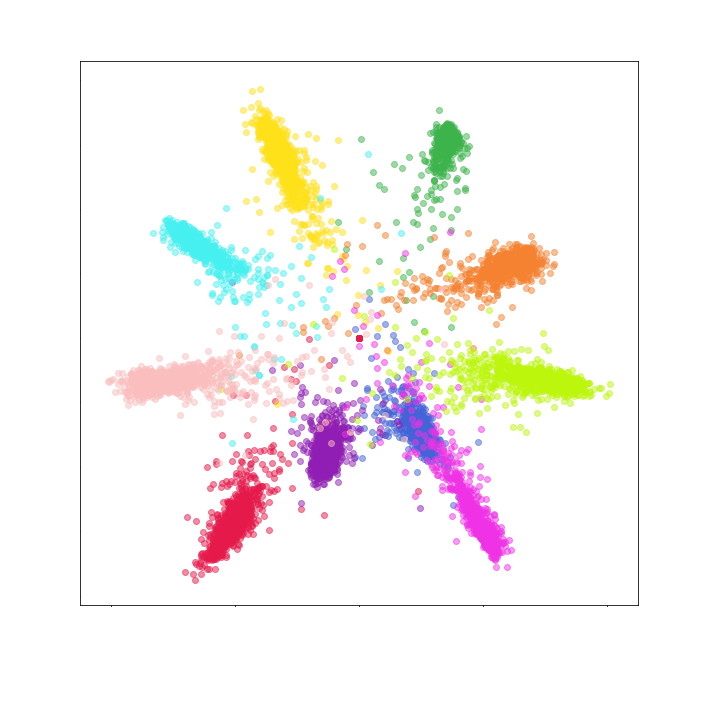}
  \end{minipage}   
     
   \begin{minipage}[b]{0.19\linewidth}
    \centering
    \includegraphics[width=1.0\linewidth]{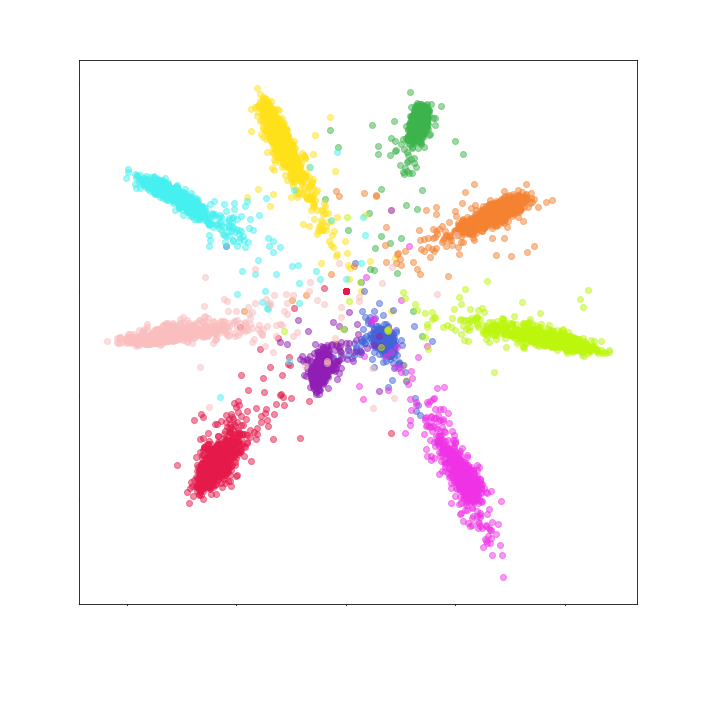}
  \end{minipage}
  \begin{minipage}[b]{0.19\linewidth}
    \centering
    \includegraphics[width=1.0\linewidth]{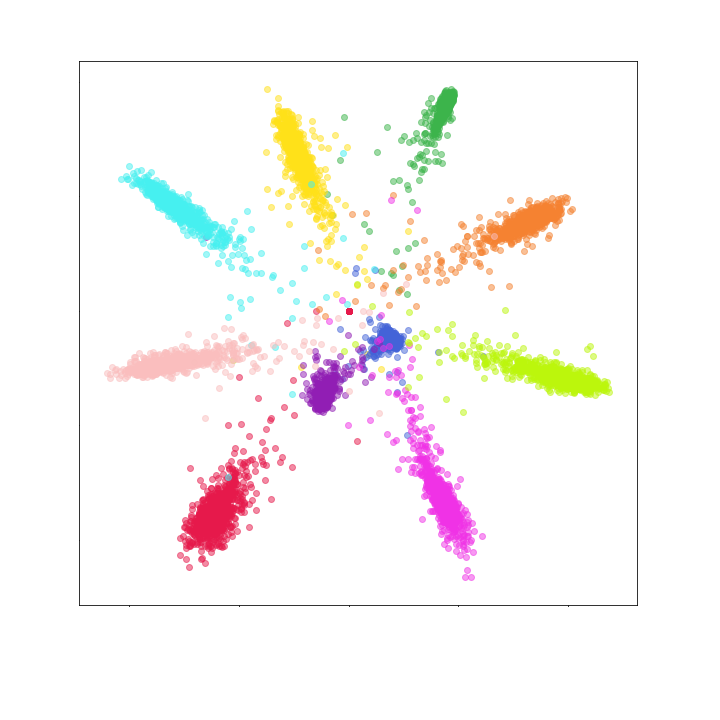}
  \end{minipage} 
  \begin{minipage}[b]{0.19\linewidth}
    \centering
    \includegraphics[width=1.0\linewidth]{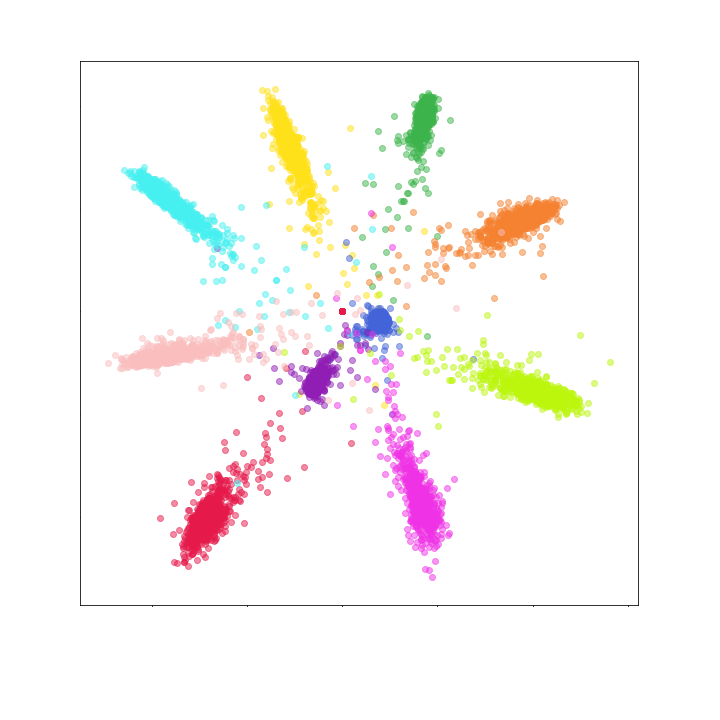}
  \end{minipage}
  \begin{minipage}[b]{0.19\linewidth}
    \centering
    \includegraphics[width=1.0\linewidth]{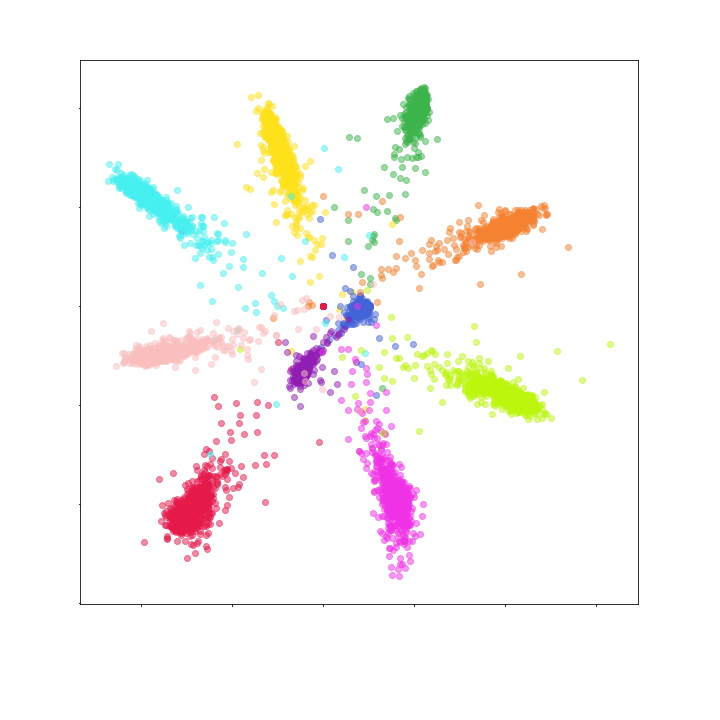}
  \end{minipage}   
  \begin{minipage}[b]{0.19\linewidth}
    \centering
    \includegraphics[width=1.0\linewidth]{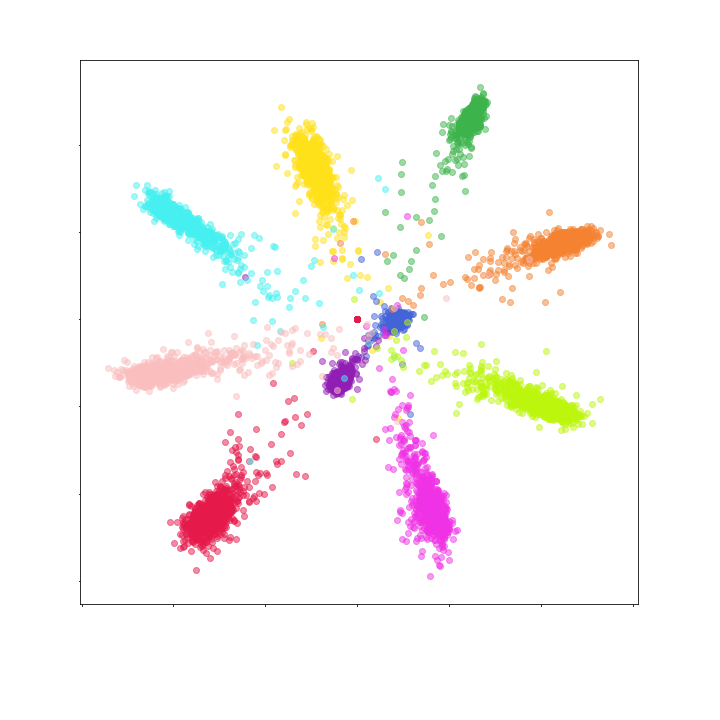}
  \end{minipage}  
  \caption{\label{step-by-step-int}
  Step-by-step visualization of centroids for 2 dimensional embeddings on MNIST dataset for intra-class variance ($\lambda= 0.05$) + cross-entropy loss function (epochs 1, 2, 3, 4, 5, 8, 11, 14, 17, 20).}

\end{figure}

\subsection{Performance metrics}
The {\em recognition  accuracy} takes the simple average success rate as the final score, counted by
\[
accuracy = \frac{\text{\it number of correct decisions}}{\text{\it total number of decisions}}
\]
We use this metric both for both, the nearest-centroid evaluation and SoftMax probability evaluation. Obviously, at testing, the decision based on  the maximum score can replace the evaluation of the SoftMax probability distribution and its maximum. Namely, we get the predicted class on test datasets by t(a) the maximum score, (b) the nearest centroid in the embedding space.  The centroids $C_k$ are learned from training data using the proposed Hadamard $\bb{H}$ layer.

%
%
%
%

\begin{figure}[htbp] 
  \begin{minipage}[b]{0.19\linewidth}
    \centering
    \includegraphics[width=1.0\linewidth]{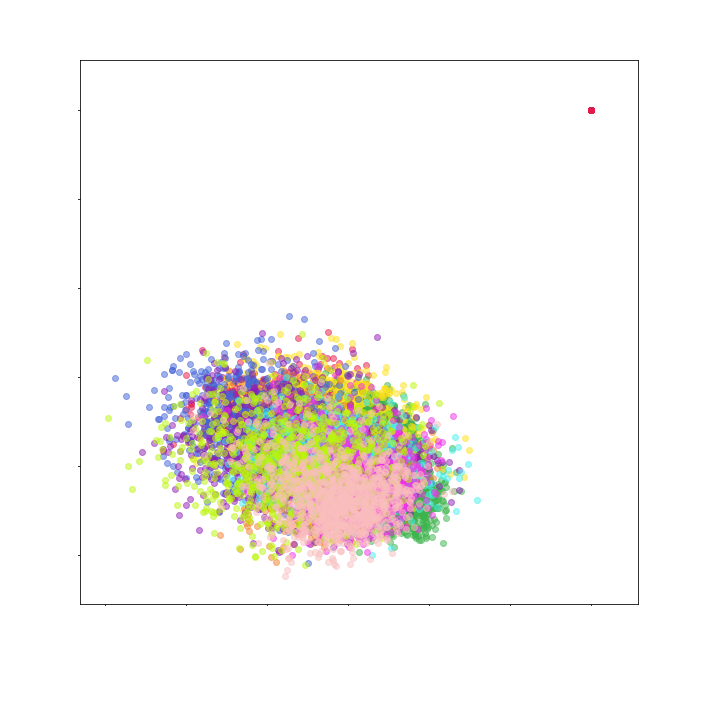}
  \end{minipage}
  \begin{minipage}[b]{0.19\linewidth}
    \centering
    \includegraphics[width=1.0\linewidth]{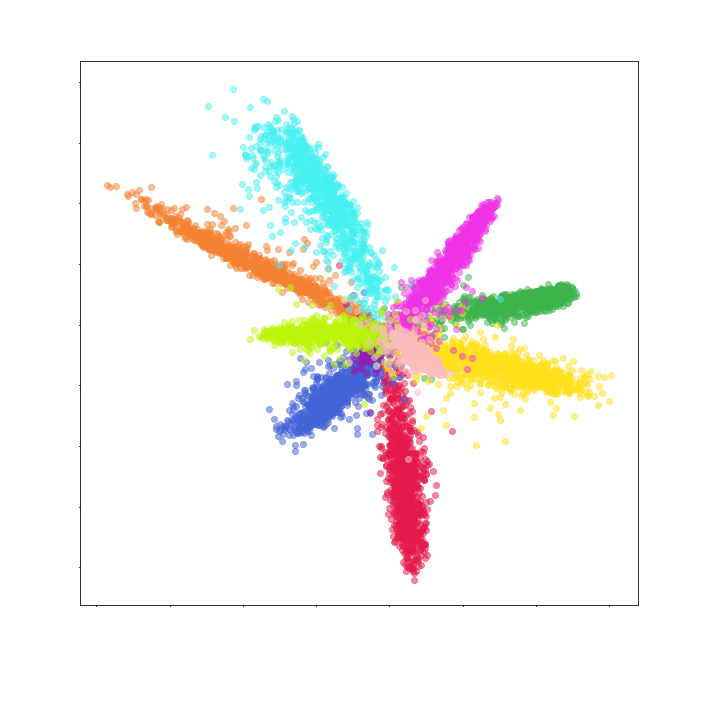}
  \end{minipage} 
  \begin{minipage}[b]{0.19\linewidth}
    \centering
    \includegraphics[width=1.0\linewidth]{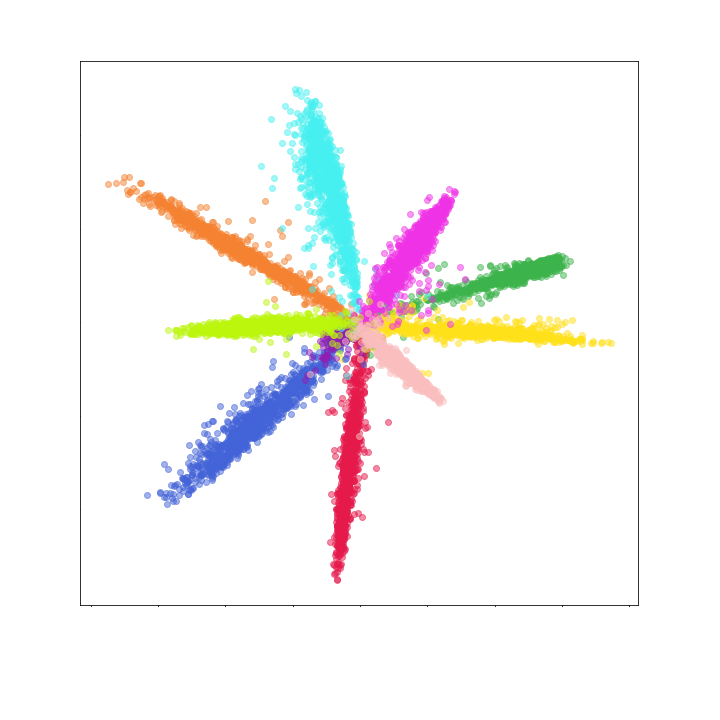}
  \end{minipage}
  \begin{minipage}[b]{0.19\linewidth}
    \centering
    \includegraphics[width=1.0\linewidth]{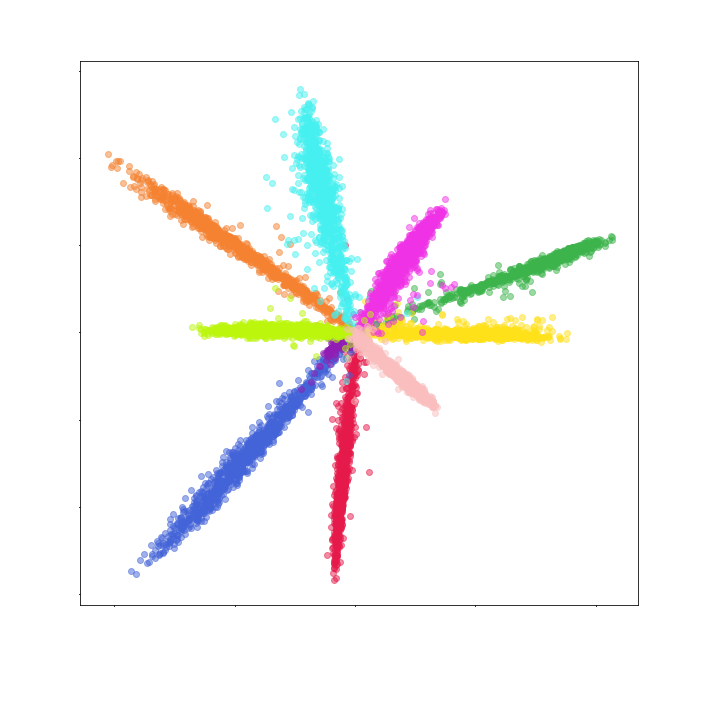}
  \end{minipage}   
  \begin{minipage}[b]{0.19\linewidth}
    \centering
    \includegraphics[width=1.0\linewidth]{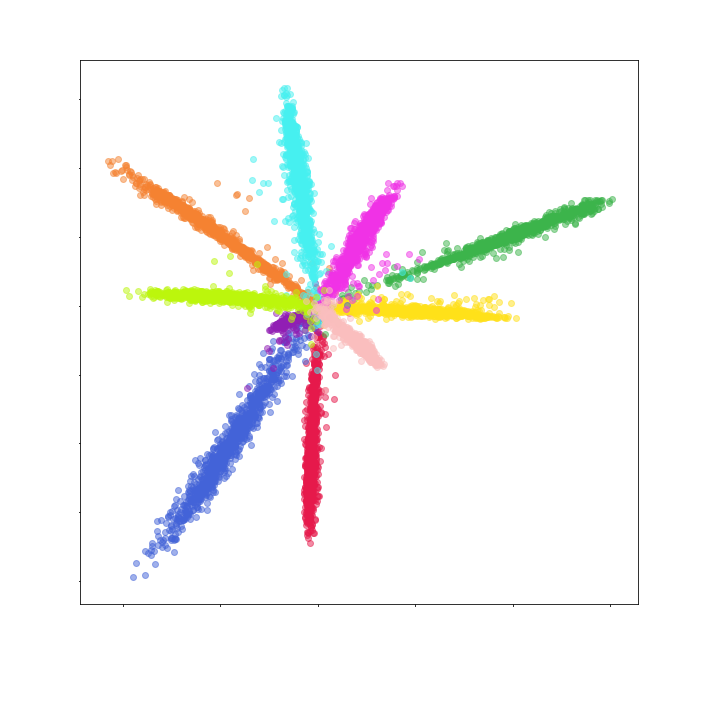}
  \end{minipage}   
     
   \begin{minipage}[b]{0.19\linewidth}
    \centering
    \includegraphics[width=1.0\linewidth]{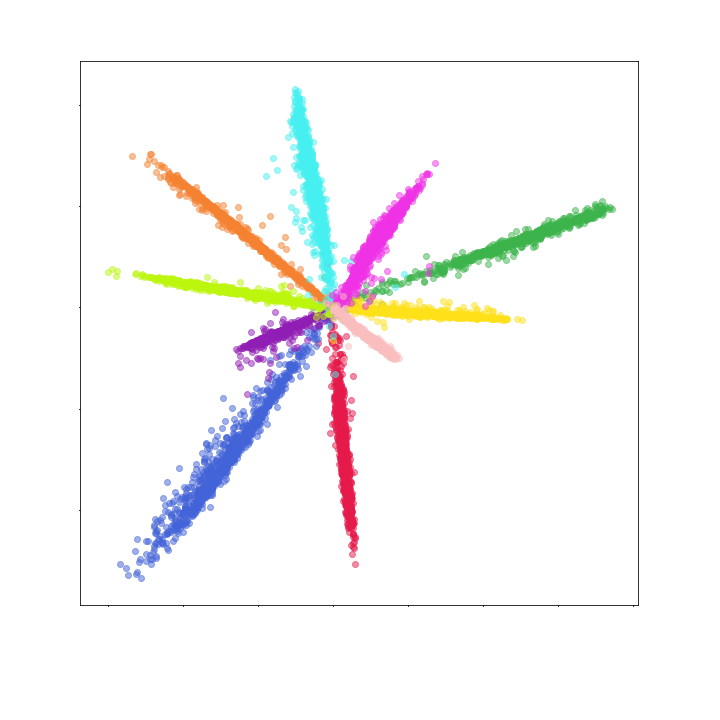}
  \end{minipage}
  \begin{minipage}[b]{0.19\linewidth}
    \centering
    \includegraphics[width=1.0\linewidth]{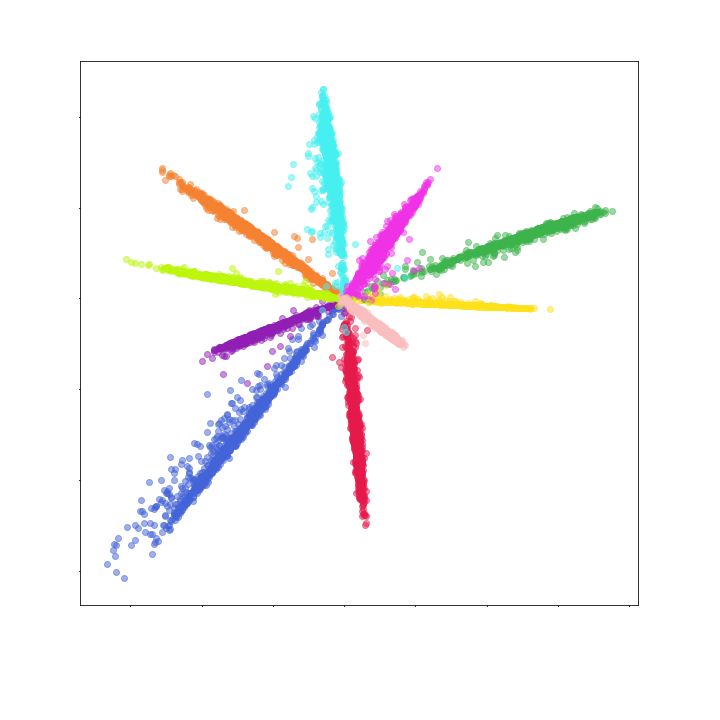}
  \end{minipage} 
  \begin{minipage}[b]{0.19\linewidth}
    \centering
    \includegraphics[width=1.0\linewidth]{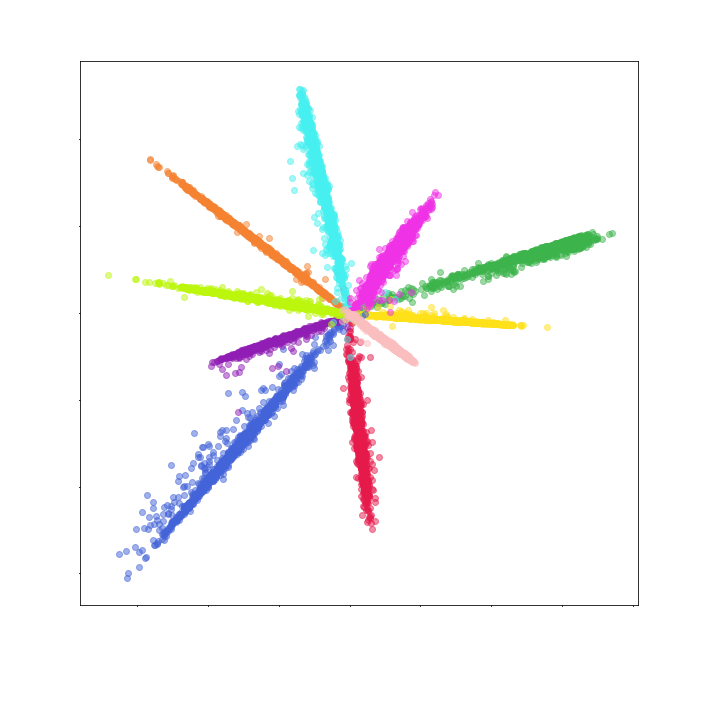}
  \end{minipage}
  \begin{minipage}[b]{0.19\linewidth}
    \centering
    \includegraphics[width=1.0\linewidth]{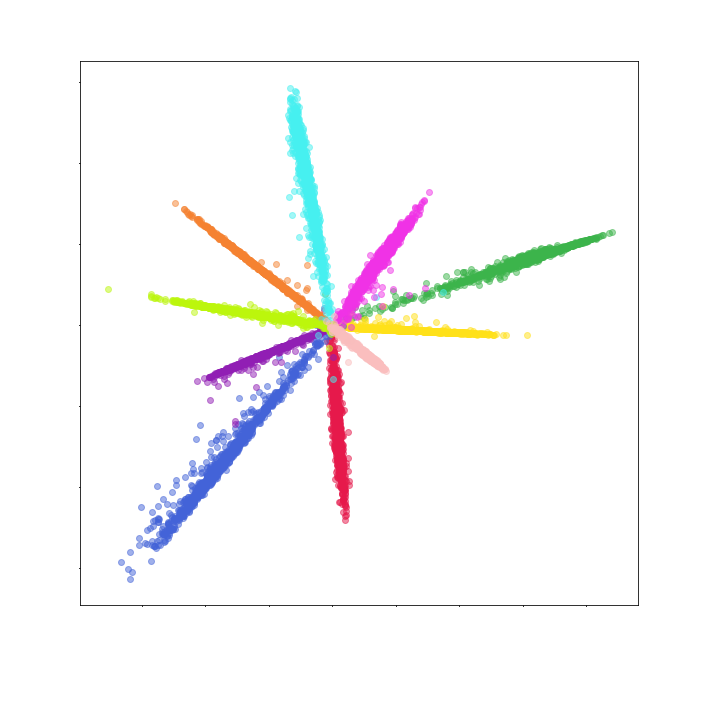}
  \end{minipage}   
  \begin{minipage}[b]{0.19\linewidth}
    \centering
    \includegraphics[width=1.0\linewidth]{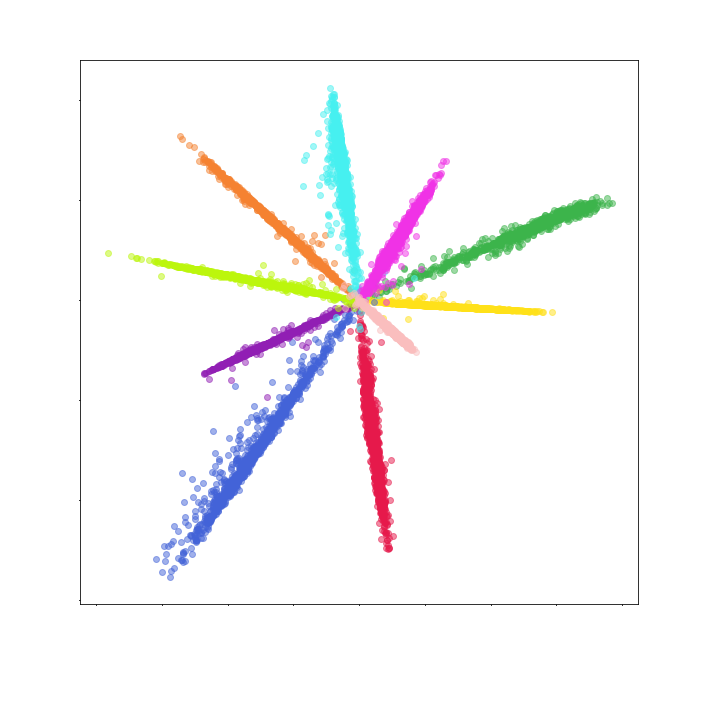}
  \end{minipage}  
  \caption{\label{step-by-step-soft} Step-by-step visualization of centroids for 2 dimensional embeddings on MNIST dataset for cross-entropy loss function (epochs 1, 2, 3, 4, 5, 8, 11, 14, 17, 20).}
\end{figure}

\subsection{Results}

We prepare visualization for two and four dimensional embeddings\footnote{For each row of presented embedding visualisation (Figure \ref{mnist-2}, \ref{mnist-4}, \ref{fashion-2},  \ref{fashion-4}) we have the following order: non-normalized SoftMax embeddings, non-normalized intra-class variance loss embeddings, normalized SoftMax embeddings and normalized intra-class variance embeddings.} to show an impact of intra-class variance loss on centroids for two-dimensional embeddings on MNIST dataset (Figure \ref{mnist-2}), Fashion-MNIST (Figure \ref{fashion-2}) and also for 4 dimensional space embeddings on both datasets (Figure \ref{mnist-4}, \ref{fashion-4}).

We can visually notice the impact of intra-class variance loss on overall distribution of embeddings (Figure  \ref{step-by-step-soft}). \textit{Long-tailed} embeddings obtained from entropy loss are packed into small clusters (Figure  \ref{step-by-step-int}).

The metrics evaluation is carried out on test datasets both for nearest centroid (Table \ref{results-centroid}) and SoftMax probability (Table \ref{results-softmax}) metrics. We observe that SoftMax probability gives similar performance on the same architecture both for the normalized and the non-normalized cases.

Results (Table \ref{results-centroid}) shows that using non-normalized features intra-class variance loss outperforms cross entropy loss for 2-dimensional and 4-dimensional embedding vectors \footnote{indicated by \textbf{bold} font}. Using smaller $\mathbf{\lambda}$ allows to achieve better results in nearest centroid metric. In addition, the introduction of an additional loss function component does not worsen the results of maximum score metric (Table \ref{results-softmax}). 

Presented loss function and Hadamard layer should be tested on more difficult multi-class problems and larger training sets. Searching hyperparameter space for complex computational algorithms is expensive, so our example can be used as a starting point. This is undoubtedly the disadvantage of each algorithm - selection: how to choose size of the embedding vector and $\mathbf{\lambda_i}$ coefficients for each component of the loss function.

\begin{table}[htbp]
\centering
\caption{\label{results-centroid}Results for minimum distance to the centroid (class mean).}
\begin{tabular}{|l|l|l|l|l|l|}
\hline
\multirow{2}{*}{} & \multicolumn{4}{c|}{\textbf{Nearest Centroid}} & \textbf{} \\ \cline{2-6} 
 & \multicolumn{2}{c|}{\textbf{Non-Norm.}} & \multicolumn{2}{c|}{\textbf{Norm.}} &  \\ \hline
Emb. Dim. & 2x1 & 4x1 & 2x1 & 4x1 &  \\ \hline
\multicolumn{5}{|c|}{\textbf{Cross entropy loss}} &  \\ \hline
MNIST & 0.909 & 0.975 & 0.984& 0.988 &  \\ \hline
Fashion-MNIST & 0.746 & 0.869 & 0.814 & 0.894 &  \\ \hline
\multicolumn{5}{|c|}{\textbf{Intra-class variance}} & $\mathbf{\lambda}$ \\ \hline
MNIST & 0.973 & 0.992 & 0.980 & 0.988 & \multirow{2}{*}{1.0} \\ \cline{1-5}
Fashion-MNIST & \textbf{0.880} & 0.891 & 0.715 & 0.884 &  \\ \hline
MNIST & 0.971 & 0.991 & \textbf{0.985} & 0.988 & \multirow{2}{*}{0.5} \\ \cline{1-5}
Fashion-MNIST & 0.870 & 0.902 & 0.738 & 0.889 &  \\ \hline
MNIST & 0.972 & \textbf{0.992} & 0.984 & 0.990 & \multirow{2}{*}{0.05} \\ \cline{1-5}
Fashion-MNIST & 0.867 & \textbf{0.910} & 0.820 & 0.891 &  \\ \hline
\end{tabular}
\end{table}

\begin{table}[htbp]
\centering
\caption{\label{results-softmax}Results for maximum score (softmax).}
\begin{tabular}{|l|l|l|l|l|l|}
\hline
\multirow{2}{*}{} & \multicolumn{4}{c|}{\textbf{Highest Score}} & \textbf{} \\ \cline{2-6} 
 & \multicolumn{2}{c|}{\textbf{Non-Norm.}} & \multicolumn{2}{c|}{\textbf{Norm.}} &  \\ \hline
Emb. Dim. & 2x1 & 4x1 & 2x1 & 4x1 &  \\ \hline
\multicolumn{5}{|c|}{\textbf{Shannon information}} &  \\ \hline
MNIST & 0.986 & 0.994 & 0.986 & 0.994 &  \\ \hline
Fashion-MNIST & 0.895 & 0.924 & 0.894 & 0.920 &  \\ \hline
\multicolumn{5}{|c|}{\textbf{intra-class variance}} & $\mathbf{\lambda}$ \\ \hline
MNIST & 0.985 & 0.994 & 0.981 & 0.995 & \multirow{2}{*}{1.0} \\ \cline{1-5}
Fashion-MNIST & 0.882 & 0.904 & 0.884 & 0.915 &  \\ \hline
MNIST & 0.986 & 0.995 & 0.982 & 0.995 & \multirow{2}{*}{0.5} \\ \cline{1-5}
Fashion-MNIST & 0.890 & 0.918 & 0.889 & 0.913 &  \\ \hline
MNIST & 0.986 & 0.995 & 0.985 & 0.994 & \multirow{2}{*}{0.05} \\ \cline{1-5}
Fashion-MNIST & 0.895 & 0.927 & 0.891 & 0.926 &  \\ \hline
\end{tabular}
\end{table}

\section*{Conclusion}

We conducted many experiments using convolutional neural network and proposed contribution to intra-class variance minimization using standard optimization techniques through Hadamard $\bb{H}$ layer. Embeddings extracted using CNN and trained using intra-class variance are less sparse than embeddings extracted directly from classifier trained with entropy objective function.\\
Intra-class  variance enables to get better results for nearest-centroid evaluation for both test dataset and embedding dimensions (Table  \ref{results-centroid}):
\begin{enumerate}
    \item 2-dim: \textbf{0.880}, \textbf{0.985} 
    \item 4-dim: \textbf{0.992}, \textbf{0.910}
\end{enumerate}

We observed that for embeddings the normalization in Euclidean norm, i.e. the projection on to the unit sphere, is not effective for intra-class variance objective. The Shannon information based embeddings exhibit better results only after normalization, however for non-normalized embeddings learned by intra-class variance the success rate is higher .

This work shows potential usability of intra-class variance loss (or regularization term) for classifiers that can be trained in supervised fashion without any additional arbitrary strategy for selection of training examples or on-the-fly statistics calculations, as you can find for works referring to triplet-loss \cite{schroff_facenet:_2015} or contrastive-loss \cite{Hadsell:2006:DRL:1153171.1153654}.

\section*{Acknowledgment}
The presented research was conducted by both authors within  the Statutory Works of Faculty of Electronics and Computer Technology, Warsaw University of  Technology, Poland.

\bibliographystyle{fcds_abbrv}
\bibliography{citations}
\end{document}